\documentclass{iopjournal}

\usepackage[square,numbers,sort&compress]{natbib}
\usepackage{mathtools, amsmath, amssymb, amsthm}
\usepackage{subcaption}
\usepackage[version=4]{mhchem} %
\usepackage{booktabs} %
\usepackage{multirow} %
\usepackage{xcolor} 

\usepackage{hyperref}       %
\usepackage{url}            %
\usepackage{hypernat}
\newtheorem{theorem}{Theorem}
\usepackage{etoc}
\usepackage{bm}
\usepackage[resetlabels]{multibib}
\newcites{si}{Supplementary References}
\usepackage[normalem]{ulem}

\newcommand{\ColorUl}[2]{%
  \bgroup
  \sbox0{#2}%
  \makebox[\wd0][c]{%
    \makebox[0pt][c]{\raisebox{-3pt}[0pt][0pt]{\color[RGB]{#1}\rule{\dimexpr\wd0-1.5pt\relax}{2pt}}}%
    \makebox[0pt][c]{\textcolor{black}{#2}}%
  }%
  \egroup
}
\usepackage{layouts}

\renewcommand{\articletype}[1]{%
  {\noindent\Large\sffamily Accepted Manuscript\par}
  \vspace{3mm}
  {\noindent\small
    \noindent This is the Accepted Manuscript version of an article accepted for publication in IOP Machine Learning: Science and Technology. IOP Publishing Ltd is not responsible for any errors or omissions in this version of the manuscript or any version derived from it. This Accepted Manuscript is published under a CC BY licence. The Version of Record is available online at 10.1088/2632-2153/ae7d85.\par
  }
  \vspace{4mm}
  {\noindent\scriptsize\sffamily
    \bfseries\MakeUppercase{#1}\par}
  \vspace{2mm}
}

\begin{document}

\etocdepthtag.toc{mt}

\articletype{Paper} %

\title{Continuous SUN (Stable, Unique, and Novel) Metric for Generative Modeling of Inorganic Crystals}

\author{Masahiro Negishi$^1$\orcid{0009-0000-2003-0256}, Hyunsoo Park$^1$\orcid{0000-0001-9388-173X}, Kinga Oliwia Mastej$^1$\orcid{0009-0006-5656-6646}, and Aron Walsh$^{1,*}$\orcid{0000-0001-5460-7033}}

\affil{$^1$Department of Materials, Imperial College London, London, United Kingdom}

\affil{$^*$Author to whom any correspondence should be addressed.}

\email{m.negishi25@imperial.ac.uk, a.walsh@imperial.ac.uk}


\begin{abstract}
To address pressing scientific challenges such as climate change, increasingly sophisticated generative models are being developed to efficiently sample the large chemical space of potential functional materials. The proliferation of these models has necessitated the establishment of rigorous evaluation metrics. While uniqueness (U), novelty (N), and stability (S) of samples serve as standard metrics, their current formulations show several limitations. U and N rely on binary comparisons of crystals, rendering them dependent on heuristic thresholds, incapable of quantifying the degree of similarity, sensitive to atomic coordinate perturbations, and not invariant to sample permutation. Similarly, the binary assessment of S risks a premature exclusion of marginally unstable yet potentially novel candidates. These limitations are addressed by making the aforementioned metrics continuous. Furthermore, we integrate them into a unified metric ``continuous SUN" (cSUN), which offers a smoother score distribution and greater tunability than the conventional binary SUN metric. Experimental results demonstrate that our continuous metrics provide granular insights into sample distributions and secondarily serve as a soft screening score for selecting samples for further analysis. Finally, the use of cSUN as a reward signal in reinforcement learning is explored, showing that its adjustable weighting scheme effectively enhances sample diversity and convergence to a better optima in a Chemeleon2 case study.
\end{abstract}

\section{Introduction}
One of the greatest fundamental challenges in materials science is the efficient design of functional crystals from a vast space of candidate chemical compositions and three-dimensional structures. 
Traditionally, researchers have relied heavily on their chemical knowledge to select promising candidates, followed by experimental or computational validation. 
However, such approaches are often prohibitively inefficient in addressing pressing energy and environmental challenges such as climate change \citep{rolnick2022tackling}.
Recent years have witnessed a growing interest in machine learning (ML) generative models, which can rapidly sample numerous candidate compounds by learning from large databases of known crystals \cite{park2024has,cheng2025ai,handoko2025artificial,de2025review,li2025materials}.

\begin{figure}
    \centering
    \includegraphics[width=\linewidth]{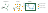}
    \caption{Illustration of the three primary evaluation metrics: Uniqueness, Novelty, and Stability. Uniqueness quantifies the internal diversity of generated samples, while novelty assesses their dissimilarity from the training samples. Both metrics fundamentally rely on the distance function $d$ to compare two crystals. Stability evaluates thermodynamic plausibility by calculating the energy above the convex hull relative to a reference database of known materials.}
    \label{fig:introduction}
\end{figure}

The rapid increase in generative models for crystals has highlighted the need for rigorous evaluation metrics \citep{duval2025lemat, yan2025mgb, hagemann2025transport}.  
Such metrics play a vital role in identifying promising models for further development and guiding scientists towards the most suitable models for their specific applications.
The performance of generative models has typically been assessed based on the uniqueness (U), novelty (N), and stability (S) of the generated samples, as well as their combination (SUN) \citep{MatterGen2025, de2025review}.
Specifically, U measures the diversity within a set of generated samples, N quantifies dissimilarity from the training data, and S evaluates physical plausibility via energy calculation (Figure~\ref{fig:introduction}).
However, existing definitions of these metrics are limited in a number of ways.
First, U and N typically rely on the ``fit” method of the StructureMatcher class in the pymatgen library to determine whether two structures are identical \citep{ong2013python}. 
The binary nature of this approach means that it is heavily dependent on heuristic thresholds and fails to quantify the degree of similarity.
In addition, small perturbations of atomic positions may alter the matching result, thereby hindering robust practical assessments \citep{widdowson2022resolving}. 
Furthermore, we demonstrate that the average U score of a set of generated samples is not invariant to permutations of the samples.
The stability evaluation (S) is also usually conducted in a binary way. 
Crystals with energy above the convex hull ($E_\text{hull}$) below the predefined threshold (e.g. 0.1 [eV/atom]) are considered (meta)stable. 
This approach fails to distinguish between structures just above the threshold and those that are highly unstable. 
Consequently, novel and interesting crystals with $E_\text{hull}$ slightly higher than the threshold have no contribution to the final model evaluation score under the current protocol.

To address these limitations, this study proposes a method for continuously measuring U, N, and S.
Specifically, we replace the discrete distance behind U and N with a combination of two continuous distances between crystals (compositional and structural), thereby capturing subtle smooth variations in crystal structures more effectively.
Additionally, we introduce a continuous stability metric based on a monotonically decreasing, continuous function of $E_\mathrm{hull}$. 
By integrating these continuous metrics, we propose the continuous SUN (cSUN) metric. 
Unlike the conventional binary SUN, our cSUN provides a smoother score distribution, allowing each sample to contribute to the final model score in proportion to its quality.
Another advantage is the inclusion of tunable weights, which allow us to prioritize one component (S, U, or N) over the others.
Experimental results demonstrate that our continuous metrics provide granular insights into the distribution of generated samples and, when used as soft screening scores, help identify samples for further analysis.

Finally, we also investigate the potential of using reinforcement learning (RL) to directly steer generative models towards producing samples with high (c)SUN scores.
Given that evaluation metrics essentially define the desired characteristics of the crystals, they are well-suited to serve as a reward function in RL.
Our results indicate that the adjustability of the cSUN weights is effective in enhancing uniqueness of samples and convergence to a better local optima in the tested Chemeleon2 setup \citep{park2025guiding}.

\section{Uniqueness and Novelty}
\label{sec:un}
We begin by focusing on the uniqueness and novelty. 
Given the lack of a universally accepted definition for these terms, Section~\ref{sec:sec:un_definition} first establishes the definitions adopted in this study.
Our definitions unify various existing metrics from the perspective of the underlying distance functions used to measure crystal similarity. 
Subsequently, Section~\ref{sec:sec:distance_definition} introduces the various distance measures employed to define these metrics.
Then, Section~\ref{sec:sec:distance_theory} provides a theoretical analysis contrasting the inherent limitations of conventional binary distances with the distinct advantages of our continuous counterparts.
Finally, Section~\ref{sec:sec:un_experiment} demonstrates the importance of carefully selecting an underlying distance function through extensive experiments with representative generative models.

\subsection{Definitions}
\label{sec:sec:un_definition}

\paragraph{Uniqueness}
A fundamental requirement for generative models is the capacity to produce diverse and non-redundant samples.
The ``uniqueness” of the generated samples quantifies this property.
Consider a set of generated crystals $X \coloneqq \{x_1, x_2, \ldots, x_n\}$.
(Strictly speaking, $X$ is a multiset allowing for duplicated samples, but for the sake of simplicity, we refer to it as a set.)
Then, the uniqueness of an individual sample $x_i$ can be formulated using either a discrete or continuous approach:
\begin{equation}
    \label{eq:sample-level-u}
    \begin{split}
        \mathrm{U}(x_i) &\coloneqq I\left(\overset{i-1}{\underset{j=1}{\land}}\left(d_\mathrm{discrete}(x_i, x_j) \neq 0\right)\right) \in \{0, 1\},\\
        \mathrm{cU}(x_i) &\coloneqq \frac{1}{n-1} \sum_{\substack{j=1 \\ j \neq i}}^{n} d_\mathrm{continuous}(x_i, x_j) \in [0, 1],\\
    \end{split}
\end{equation}
where $d_\mathrm{discrete}$ represents an arbitrary discrete distance function (0 or 1), $I$ is an indicator function that returns 1 if the proposition is true and 0 otherwise, and $d_\mathrm{continuous}$ is an arbitrary real-valued distance function normalized to $[0, 1]$.
Technically, $d_\mathrm{discrete}$ and $d_\mathrm{continuous}$ constitute pseudometrics rather than true distances, as distinct crystals can be considered equivalent.
However, we refer to them ``distances” for simplicity.
Note that in both U and cU, a higher score indicates greater dissimilarity between $x_i$ and the other samples.
Building on this sample-level definition, the uniqueness score of a generative model is defined as the mean uniqueness score of its generated candidates:
\begin{equation*}
    \overline{\mathrm{U}} \coloneqq \begin{cases}
        \frac{1}{n} \sum_{i=1}^{n} \mathrm{U}(x_i) & \text{for discrete distances}\\
        \frac{1}{n} \sum_{i=1}^{n} \mathrm{cU}(x_i) & \text{for continuous distances}.\\
    \end{cases}
\end{equation*}
When continuous distance is employed, $\overline{\mathrm{U}}$ is equivalent to the average pairwise distance.
Prior research \citep{MatterGen2025, miller2024flowmm, levy2025symmcd, kazeev2025wyckoff, sriram2024flowllm, de2026generative, joshi2025allatom, cornet2025kinetic, hollmer2025open, park2025guiding, xu2025plaid++, antunes2024crystal, yan2024invariant, gan2025matllmsearchcrystalstructurediscovery} has typically used U and $\overline{\mathrm{U}}$ based on the discrete distance $d_\mathrm{smat}$ defined by the StructureMatcher class of pymatgen.
In Section~\ref{sec:sec:distance_theory}, we theoretically show the limitations of $d_\mathrm{smat}$ and propose using continuous distances instead.
For the remainder of this text, we adopt the notation $\mathrm{U}_\mathrm{name}$ (or $\mathrm{cU}_\mathrm{name}$) and $\overline{\mathrm{U}}_\mathrm{name}$ to denote metrics based on $d_\mathrm{name}$ (e.g., $\mathrm{U}_\mathrm{smat}$ and $\overline{\mathrm{U}}_\mathrm{smat}$).

\paragraph{Novelty}
Relying solely on uniqueness is inadequate for the rigorous evaluation of crystal generative models.
A model may be capable of generating a diverse set of outputs that are, in fact, only minor variations of structures present in the training set. 
This limitation necessitates the inclusion of another key metric: ``novelty.”
Novelty serves to quantify the degree of dissimilarity between a set of generated samples, $X$, and the training data $Y_\mathrm{train} \coloneqq \{y_1, y_2, \ldots y_m\}$.  
Analogous to uniqueness, sample-level novelty can be formulated in two ways:
\begin{equation}
    \label{eq:sample-level-n}
    \begin{split}
        \mathrm{N}(x_i) &\coloneqq I\left(\overset{m}{\underset{j=1}{\land}}\left(d_\mathrm{discrete}(x_i, y_j) \neq 0\right)\right) \in \{0, 1\},\\
        \mathrm{cN}(x_i) &\coloneqq \min_{j=1 \sim m} d_\mathrm{continuous}(x_i, y_j) \in [0, 1].\\
    \end{split}
\end{equation}
Here, a higher score indicates greater dissimilarity between $x_i$ and $Y_\mathrm{train}$.
The model-level novelty score is also defined in a similar manner:
\begin{equation*}
    \overline{\mathrm{N}} \coloneqq \begin{cases}
        \frac{1}{n} \sum_{i=1}^{n} \mathrm{N}(x_i) & \text{for discrete distances}\\
        \frac{1}{n} \sum_{i=1}^{n} \mathrm{cN}(x_i) & \text{for continuous distances}.\\
    \end{cases}
\end{equation*}
As with uniqueness, discrete N and $\overline{\mathrm{N}}$ based on the StructureMatcher class of pymatgen have remained the predominant choice in previous studies \citep{MatterGen2025, miller2024flowmm, levy2025symmcd, kazeev2025wyckoff, sriram2024flowllm, de2026generative, joshi2025allatom, cornet2025kinetic, hollmer2025open, park2025guiding, xu2025plaid++, antunes2024crystal, yan2024invariant, gan2025matllmsearchcrystalstructurediscovery}.
Adopting the notation introduced earlier, we denote N (or cN) and $\overline{\mathrm{N}}$ based on $d_\mathrm{name}$ as $\mathrm{N}_\mathrm{name}$ (or $\mathrm{cN}_\mathrm{name}$) and $\overline{\mathrm{N}}_\mathrm{name}$.

\subsection{Crystal Distance Functions}
\label{sec:sec:distance_definition}

\begin{figure}
    \centering
    \begin{minipage}[b]{0.19\columnwidth}
        \includegraphics[width=0.95\columnwidth]{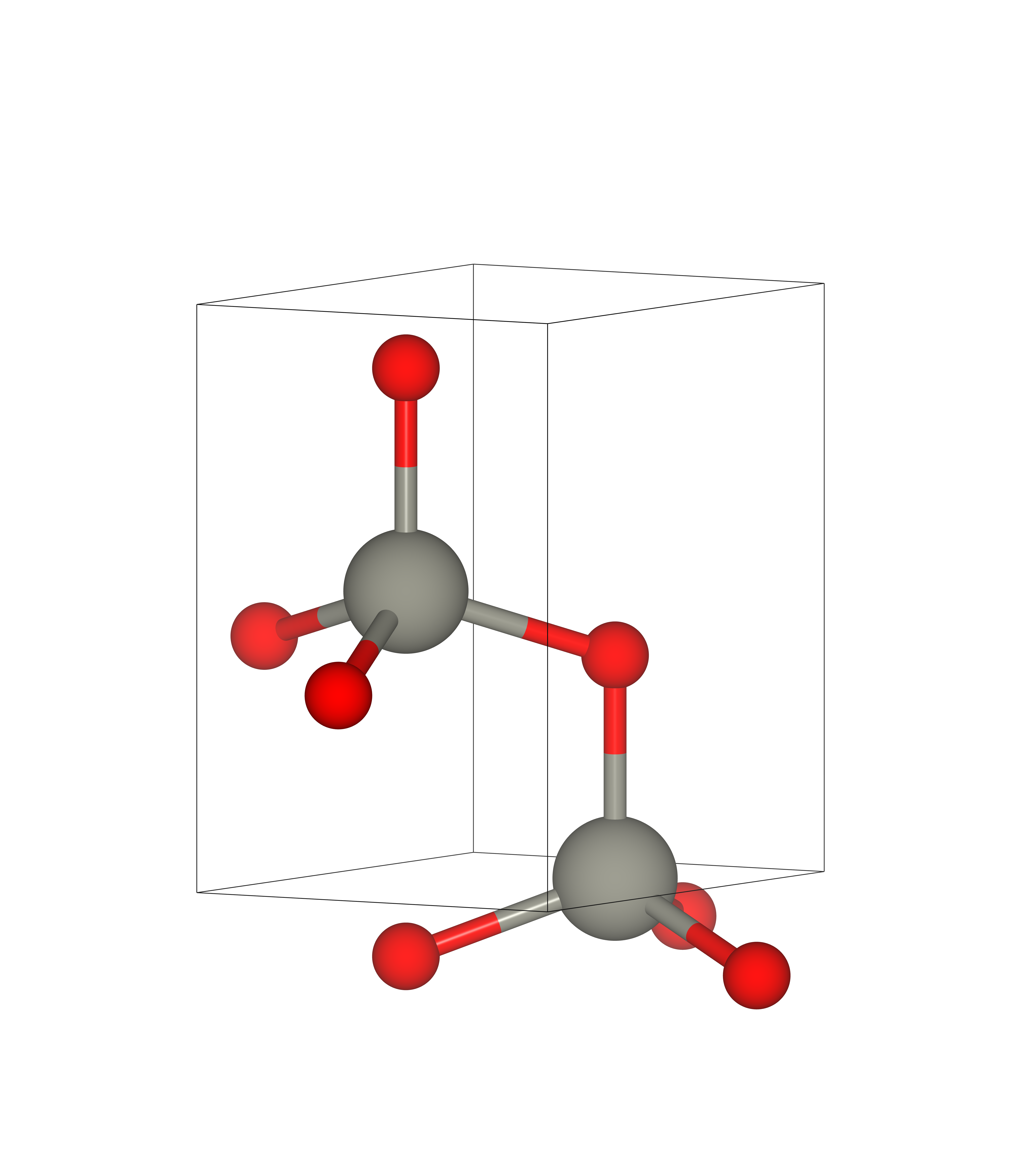}
        \subcaption{wz-ZnO (mp-2133)}
    \end{minipage}
    \begin{minipage}[b]{0.19\columnwidth}
        \includegraphics[width=0.95\columnwidth]{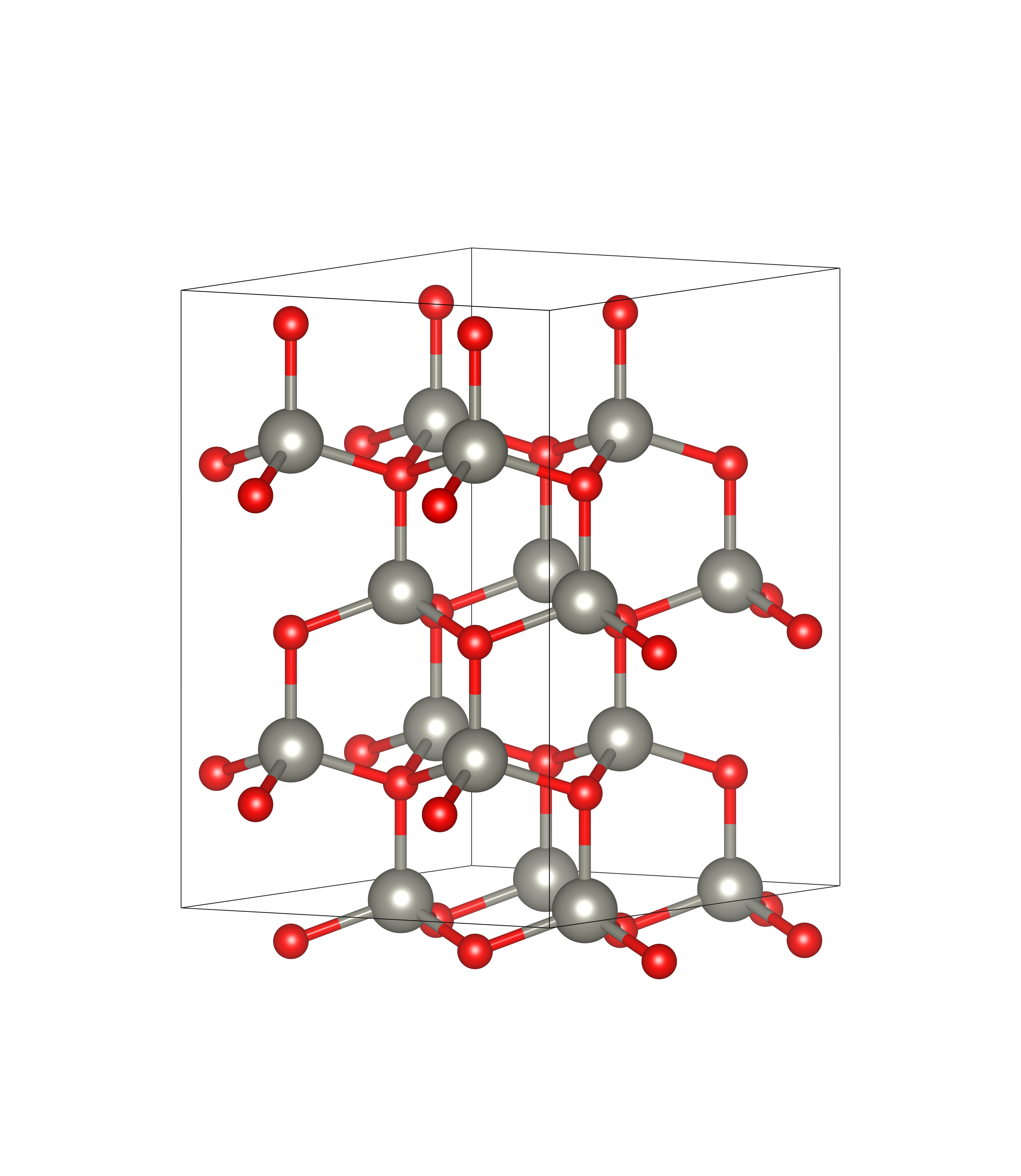}
        \subcaption{wz-ZnO* (mp-2133)}
    \end{minipage}
    \begin{minipage}[b]{0.19\columnwidth}
        \includegraphics[width=0.95\columnwidth]{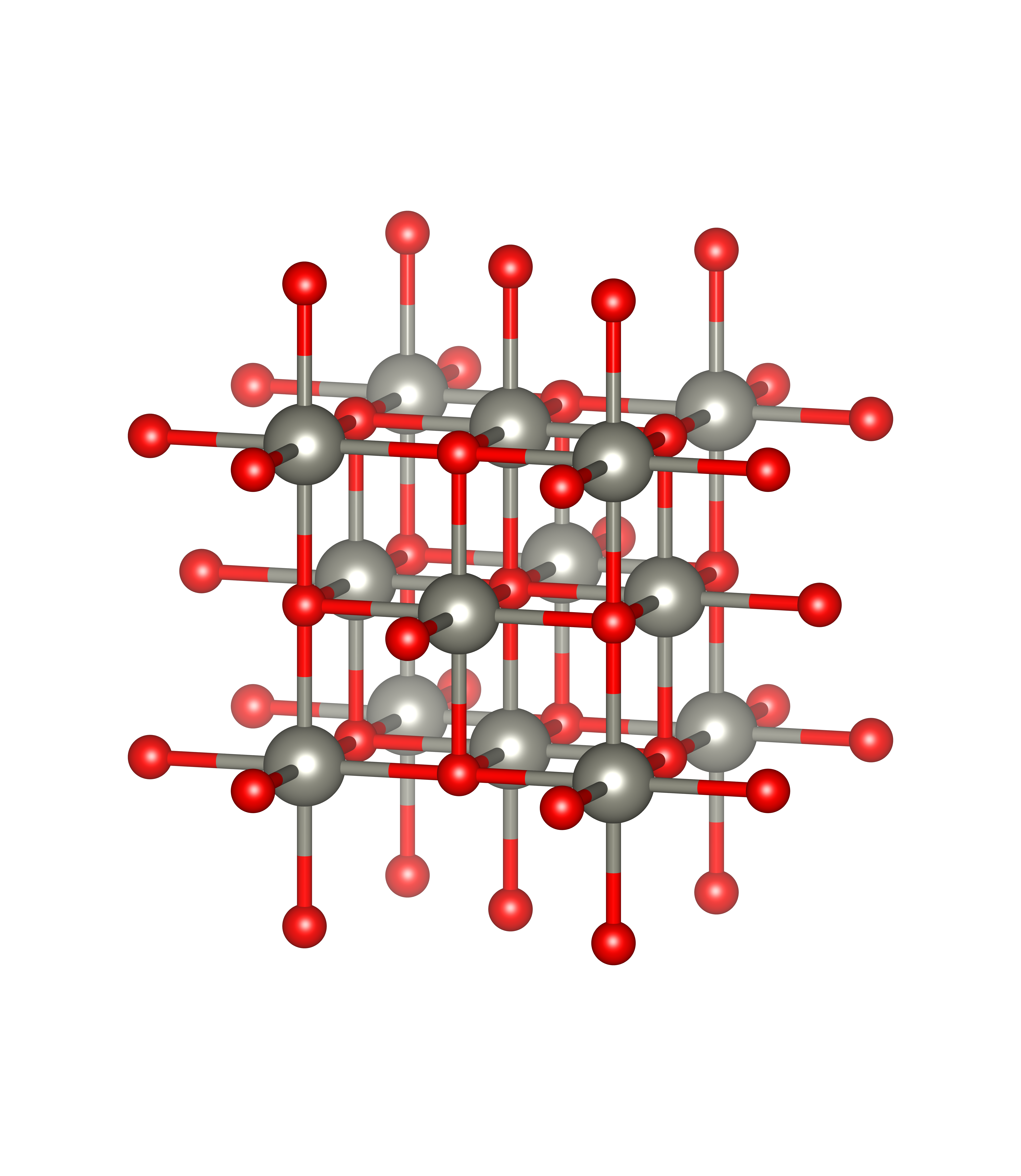}
        \subcaption{rs-ZnO (mp-2229)}
    \end{minipage}
    \begin{minipage}[b]{0.19\columnwidth}
        \includegraphics[width=0.95\columnwidth]{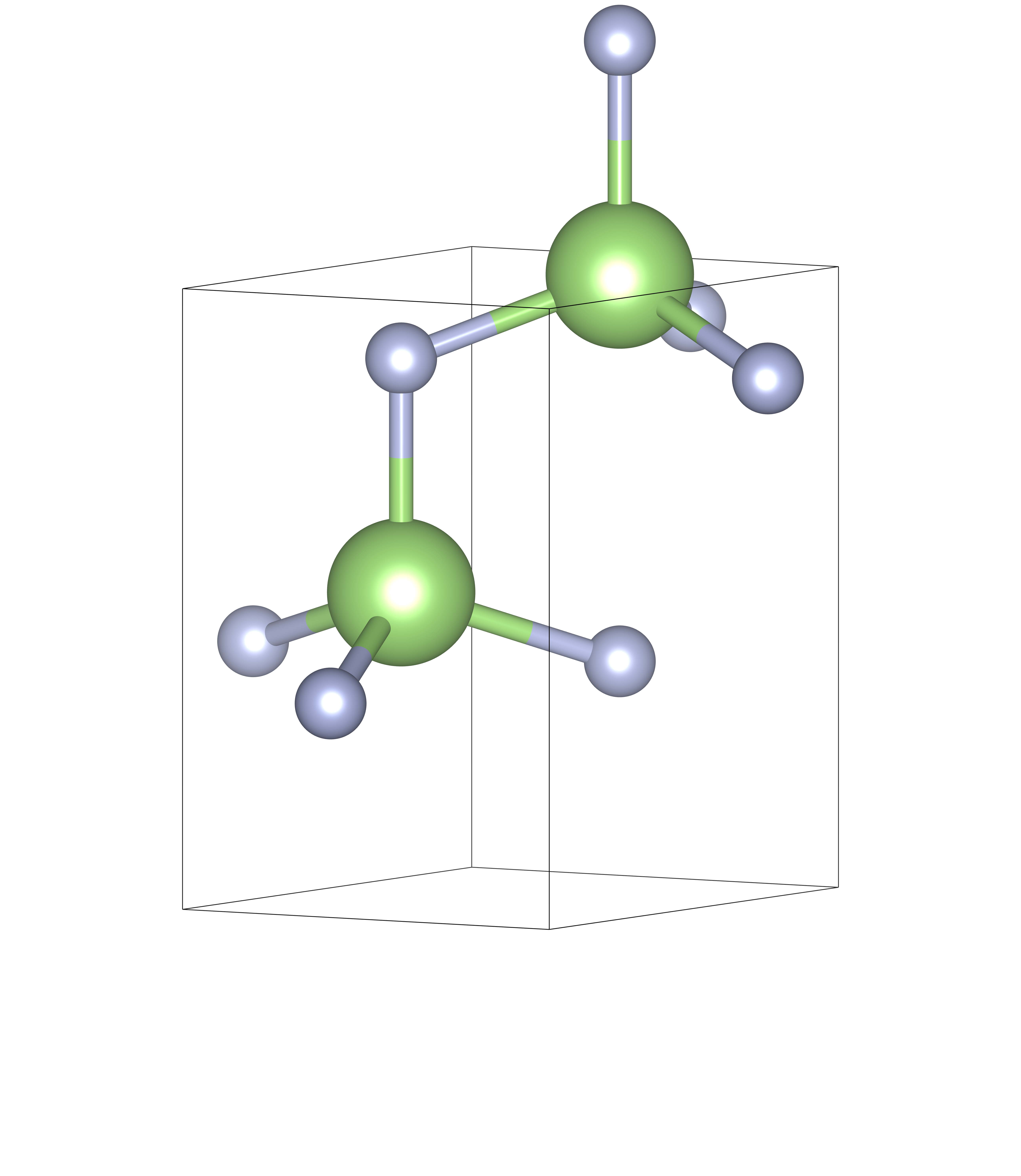}
        \subcaption{wz-GaN (mp-804)}
    \end{minipage}
    \begin{minipage}[b]{0.19\columnwidth}
        \includegraphics[width=0.95\columnwidth]{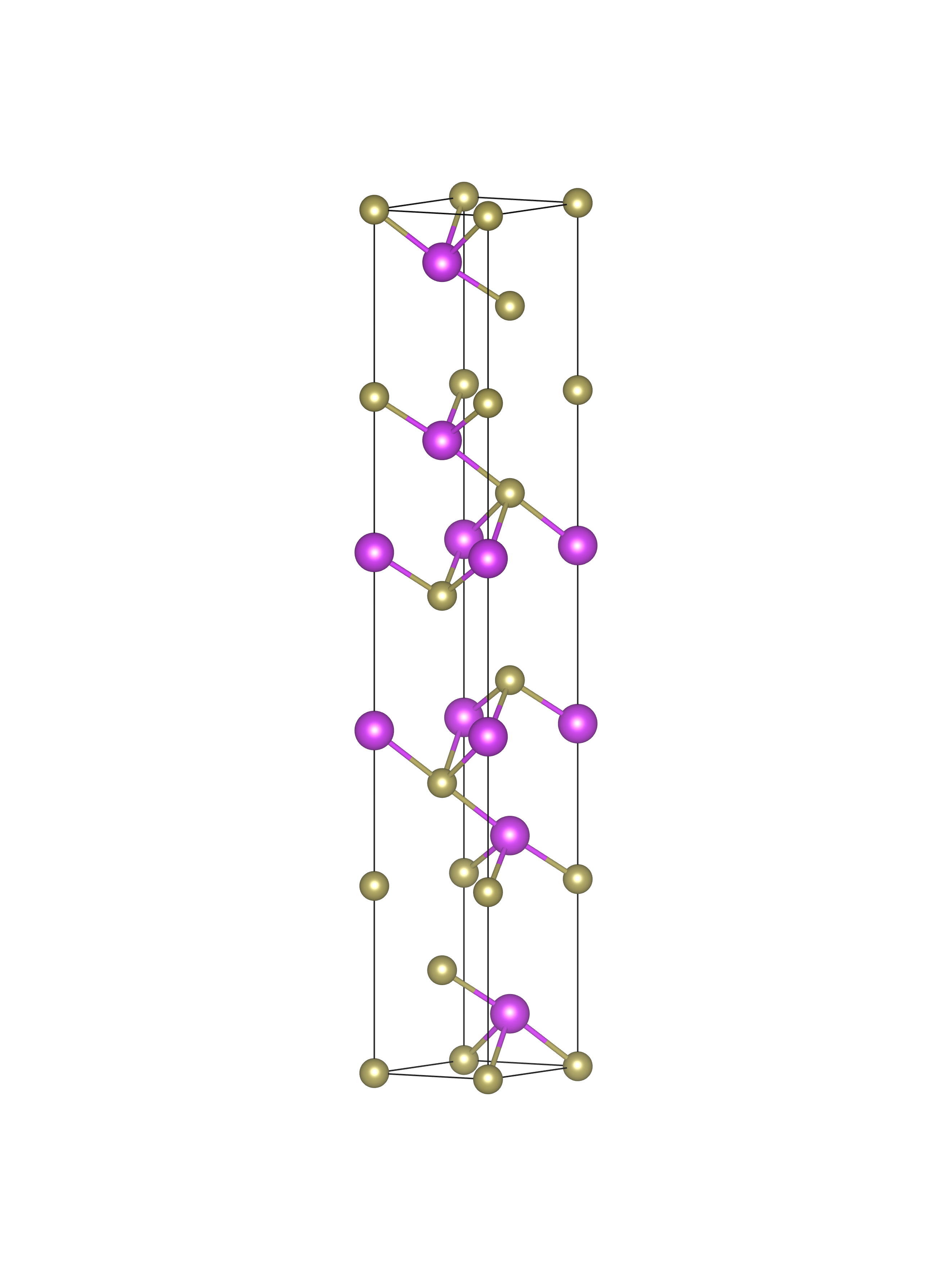}
        \subcaption{\ce{Bi2Te3} (mp-34202)}
    \end{minipage}
    \caption{Unit cell (and $2\times2\times2$ supercell*) of example crystal structures used in Table~\ref{tab:example}.}
    \label{fig:example}
\end{figure}

\begin{table}
    \caption{Distances between example pairs of crystals based on different metrics. Continuous distances ($d_\mathrm{elm}$, $d_\mathrm{am}$, and $d_\mathrm{elm{+}am}$) offer more granular insights into crystal similarity than their discrete counterparts ($d_\mathrm{comp}$, $d_\mathrm{wyckoff}$, and $d_\mathrm{smat}$).}
    \label{tab:example}
    \centering
    \begin{tabular}{@{}lcccccc@{}}
        \toprule
        & \multicolumn{2}{c}{compositional} & \multicolumn{2}{c}{structural} & \multicolumn{2}{c}{both}\\
        \cmidrule(r){2-3}\cmidrule(r){4-5}\cmidrule(r){6-7}
        & discrete & continuous & discrete & continuous & discrete & continuous \\
        \cmidrule(r){2-7}
        & $d_\mathrm{comp}$ & $d_\mathrm{elm}$ & $d_\mathrm{wyckoff}$ & $d_\mathrm{am}$ & $d_\mathrm{smat}$ & $d_\mathrm{elm{+}am}$ \\
        \midrule
        wz-ZnO vs wz-ZnO (supercell) & 0 & 0.000 & 0 & 0.000 & 0 & 0.000\\ 
        wz-ZnO vs rs-ZnO & 0 & 0.000 & 1 & 0.510 & 1 & 0.114\\ 
        wz-ZnO vs wz-GaN & 1 & 0.875 & 0 & 0.108 & 1 & 0.699\\
        wz-ZnO vs \ce{Bi2Te3} & 1 & 0.915 & 1 & 0.783 & 1 & 0.881\\ 
        \bottomrule
    \end{tabular}
\end{table}

As shown in Equations~\eqref{eq:sample-level-u} and \eqref{eq:sample-level-n}, a distance function measuring crystal similarity serves as the fundamental basis for assessing uniqueness and novelty.
This section introduces various distance functions employed in prior studies, alongside the continuous distances used in our approach.
Table~\ref{tab:example} illustrates the distinct characteristics of these metrics on representative structures from the Materials Project \citep{jain2013commentary}, visualized in Figure~\ref{fig:example}.
For detailed definitions of the distance metrics introduced herein, as well as a discussion of other metrics excluded from this work, see \ref*{appendix:sec:sec:distance_definition} and \ref*{appendix:sec:sec:other_distances} of the Supporting Information.

\paragraph{\bm{$d_\mathrm{smat}$}}
The widely adopted distance function $d_\mathrm{smat}$ based on the StructureMatcher \citep{ong2013python} operates in two steps. 
(1) It first evaluates compositional identity; distinct compositions immediately yield a distance of 1.
(2) Contingent on a compositional match, the algorithm reduces both structures to their primitive cells and explores all possible lattice transformations and translations to align one structure onto the other. 
A match is confirmed ($d_\mathrm{smat} = 0$) if the maximum Euclidean distance between corresponding atoms, minimized across all potential alignments, falls within a predefined threshold. 
In the absence of such an alignment, the function returns a distance of 1.
Regarding the examples in Table~\ref{tab:example}, $d_\mathrm{smat} = 1$ for every pair except for the first one, where two crystals are compositionally and structurally identical.

\paragraph{\bm{$d_\mathrm{comp}$} and \bm{$d_\mathrm{wyckoff}$}} 
One limitation of $d_\mathrm{smat}$ is that $d_\mathrm{smat}=1$ does not indicate whether the difference arises from composition or structure.
This ambiguity presents a challenge when either composition or structure is prioritized over the other. 
For instance, structural uniqueness and novelty may be of greater significance when searching for crystals with specific mechanical properties. 
To facilitate a more targeted assessment, specialized distance functions have been introduced. 
Regarding compositional comparison, $d_\mathrm{comp}$ returns 0 if and only if two crystals share the identical composition \citep{zhao2023physics, ren2022invertible, govindarajan2024learning, vasylenko2025physics}. 
Similarly, for structural comparison, $d_\mathrm{wyckoff}$ returns 0 if and only if the structures share the same space group and Wyckoff letters for the atomic sites in the unit cell \citep{levy2025symmcd, kazeev2025wyckoff}.
In Table~\ref{tab:example}, these two distances yield a distance of 0 to one of the three pairs with $d_\mathrm{smat} = 1$, by focusing on either composition or structure.

\paragraph{\bm{$d_\mathrm{elm}$} and \bm{$d_\mathrm{am}$}}
While $d_\mathrm{comp}$ and $d_\mathrm{wyckoff}$ allow for more targeted evaluations, they remain limited by their discrete nature, which prevents the quantification of the degree of similarity.
To overcome this drawback, we propose the adoption of continuous distance functions. 
For compositional comparison, we employ the Element Mover's Distance $d_\mathrm{elm}$, which is defined as the optimal transport cost between compositions \citep{hargreaves2020earth}. 
Specifically, it treats composition as a histogram of elements and computes the minimum cost of moving one histogram to another based on the chemical similarity \citep{glawe2016optimal} of the constituent elements.
For structural comparison, we utilize $d_\mathrm{am}$, the $L_\infty$ distance between Average Minimum Distance (AMD) vectors \citep{widdowson2022average}. 
The AMD vector is a structural fingerprint for a crystal, where AMD$[k]$ denotes the mean distance from an atom to its $k^\mathrm{th}$ nearest neighbor, averaged over all atoms in a primitive unit cell. 
The AMD vector length $k_\mathrm{max}$ is a hyperparameter. Following the original work, we use the default value $k_\mathrm{max}=100$. 
In \ref*{appendix:sec:sec:am_k}, we show that evaluation metrics computed with different values of $k_\mathrm{max}$ yield nearly identical results.
It is worth noting that previous studies used $d_\mathrm{am}$ to evaluate the novelty of a single sample \citep{widdowson2025geographic}, or to identify duplicates in crystal databases or generated samples \citep{ANOSOVA2026112108}. 
However, they did not derive a single, continuous uniqueness $\overline{\mathrm{U}}$ or novelty $\overline{\mathrm{N}}$ for a set of generated samples to compare different models or distance functions.
Although $d_\mathrm{elm}$ and $d_\mathrm{am}$ are not normalized in their original formulations, we normalize them to the range [0, 1] using the formula $d = \frac{d'}{1 + d'}$, where $d'$ is the original unnormalized distance. 
This normalization ensures that the uniqueness and novelty metrics derived from these distance functions have the same range as those based on conventional binary distance functions, while preserving the properties of a proper pseudometric. 
In \ref*{appendix:sec:sec:normalize}, we compare several normalization methods and confirm that the uniqueness and novelty evaluations are largely robust to the choice of normalization.

\paragraph{\bm{$d_\mathrm{elm{+}am}$}}
Having established continuous counterparts for the compositional $d_\mathrm{comp}$ and structural $d_\mathrm{wyckoff}$, the remaining task is to construct a continuous analog of $d_\mathrm{smat}$ capable of measuring compositional and structural differences simultaneously.
To this end, we define $d_\mathrm{elm{+}am}$ as a linear combination of $d_\mathrm{elm}$ and $d_\mathrm{am}$:
\begin{equation}
    \label{eq:d_elm+am}
    d_\mathrm{elm{+}am} \coloneqq w_\mathrm{elm} \cdot d_\mathrm{elm} + w_\mathrm{am} \cdot d_\mathrm{am} \text{, where } w_\mathrm{elm} = \frac{\sigma_\mathrm{am}}{\sigma_\mathrm{elm} + \sigma_\mathrm{am}} \text{ and } w_\mathrm{am} = \frac{\sigma_\mathrm{elm}}{\sigma_\mathrm{elm} + \sigma_\mathrm{am}}.
\end{equation}
Here, $\sigma_\mathrm{elm}$ and $\sigma_\mathrm{am}$ denote the standard deviations of the distributions of pairwise distances. 
The weights $w_\mathrm{elm}$ and $w_\mathrm{am}$ are introduced to ensure that $d_\mathrm{elm}$ and $d_\mathrm{am}$ contribute equally to the combined metric $d_\mathrm{elm{+}am}$.
In this study, we computed $\sigma_\mathrm{elm}$ and $\sigma_\mathrm{am}$ using the MP20 test data, as all our experiments had been conducted with the MP20 dataset \citep{xie2022crystal}. 
The resulting weights were $w_\mathrm{elm} \approx 0.78$ and $w_\mathrm{am} \approx 0.22$.
These values can be adjusted depending on whether the evaluation should prioritize compositional or structural similarity. In \ref*{appendix:sec:sec:elm+am_coeff}, we perform a sensitivity analysis over the full range from purely structural to purely compositional comparison.
Table~\ref{tab:example} shows that $d_\mathrm{elm}$, $d_\mathrm{am}$, and $d_\mathrm{elm{+}am}$ all successfully provide a continuous alternative to discrete distances.

\subsection{Theoretical Advantages of Continuous Distance Functions}
\label{sec:sec:distance_theory}

The advantages of $d_\mathrm{elm}$, $d_\mathrm{am}$ and $d_\mathrm{elm{+}am}$ extend beyond their real-valued nature. 
This section theoretically demonstrates that these functions satisfy three fundamental criteria essential for robust evaluation. 
Specifically, our analysis encompasses two requirements established by \citep{widdowson2022resolving}, alongside a third novel condition proposed in this work.
A summary of the theoretical analysis below is shown in Table~\ref{tab:property}.

\paragraph{Isometry Invariance}
The first prerequisite from \citep{widdowson2022resolving} is that a distance function must correctly identify identical crystals regardless of their translation or rotation. 
Failure to meet this criterion could result in a single crystal being counted multiple times during uniqueness measurement, or a generated crystal being erroneously classified as novel when it is merely a different orientation of a training set structure. 
This fundamental property is called ``isometry invariance”, and is formally defined as follows \citep{widdowson2022resolving}:
\begin{center}
    For any two isometric crystals, $x \cong x'$, the distance $d(x, x') = 0$.
\end{center}
The widely employed $d_\mathrm{smat}$ satisfies this condition by performing an exhaustive search over all possible lattice transformations and translations to align input structures. 
Likewise, the compositional distances ($d_\mathrm{comp}$ and $d_\mathrm{elm}$) are trivially isometry invariant because a crystal's composition is unaffected by isometric transformations. 
$d_\mathrm{am}$ also satisfies this property since interatomic distances do not change with the crystal's overall orientation.
It follows that $d_\mathrm{elm{+}am}$ satisfies this requirement as well, because its constituents $d_\mathrm{elm}$ and $d_\mathrm{am}$ are both invariant.
In contrast, $d_\mathrm{wyckoff}$ fails to hold this property, as Wyckoff letters depend on the choice of the unit cell's origin.
While this issue could be addressed by computing $d_\mathrm{wyckoff}$ on Niggli's reduced cells with a canonical origin, we do not apply this reduction in the present study, in order to maintain consistency with previous works that utilized $d_\mathrm{wyckoff}$ without such reduction \citep{levy2025symmcd, kazeev2025wyckoff}.

\paragraph{Lipschitz Continuity}
The second requirement from \citep{widdowson2022resolving} is that a distance function should exhibit robustness against small perturbations in atomic positions.
In practice, such perturbations can be attributed to various factors, including atomic vibrations, experimental measurement errors, and fluctuations in the generative model's outputs.
This property is mathematically formalized using the concept of Lipschitz continuity \citep{widdowson2022resolving}:
\begin{center}
    There exists a positive constant $C$ such that if $x'$ is obtained from $x$ by\\ shifting each atom by a distance of at most $\varepsilon$, then $d(x, x') \le C\varepsilon$.
\end{center}
The compositional distances, $d_\mathrm{comp}$ and $d_\mathrm{elm}$, satisfy this condition trivially.
Given that they are independent of atomic coordinates, the requirement holds true for any $C>0$. 
$d_\mathrm{am}$ is also Lipschitz continuous because the underlying interatomic distances change smoothly with atomic perturbations. 
Thus, the combined metric $d_\mathrm{elm{+}am}$ is Lipschitz continuous as well.
In contrast, the discrete structural distances fail to meet the requirement.
$d_\mathrm{smat}$ lacks this property since the distance undergoes an abrupt transition from 0 to 1 when atomic perturbations induce a change in the primitive unit cell.
Similarly, $d_\mathrm{wyckoff}$ violates Lipschitz continuity because small displacements can alter the calculated space group, thereby changing the discrete Wyckoff letters.

\begin{table}
    \caption{Summary of the theoretical properties of crystal distance functions. Continuous distances satisfy all three robustness requirements, whereas the conventional $d_\mathrm{smat}$ fails to do so.}
    \label{tab:property}
    \centering
    \begin{tabular}{@{}lcccccc@{}}
        \toprule
        & \multicolumn{2}{c}{compositional} & \multicolumn{2}{c}{structural} & \multicolumn{2}{c}{both}\\
        \cmidrule(r){2-3}\cmidrule(r){4-5}\cmidrule(r){6-7}
        & discrete & continuous & discrete & continuous & discrete & continuous \\
        \cmidrule(r){2-7}
        & $d_\mathrm{comp}$ & $d_\mathrm{elm}$ & $d_\mathrm{wyckoff}$ & $d_\mathrm{am}$ & $d_\mathrm{smat}$ & $d_\mathrm{elm{+}am}$ \\
        \midrule
        Isometry invariance & \checkmark & \checkmark && \checkmark & \checkmark & \checkmark \\ 
        Lipschitz continuity & \checkmark & \checkmark && \checkmark && \checkmark \\
        Permutation invariance of $\overline{\mathrm{U}}$ & \checkmark & \checkmark & \checkmark & \checkmark && \checkmark \\
        \bottomrule
    \end{tabular}
\end{table}

\begin{table}
    \caption{Variation of $\overline{\mathrm{U}}_\mathrm{smat}$ under sample permutation. Scores were computed on a fixed set of 10k CDVAE samples shuffled with different seeds, confirming that the metric is not invariant to sample order.}
    \label{tab:invariance}
    \centering
    \begin{tabular}{@{}lccccc@{}}
        \toprule
        seed & 0 & 1 & 2 & 3 & 4 \\
        \midrule
        $\overline{\mathrm{U}}_\mathrm{smat}$ & 0.9942 & 0.9947 & 0.9940 & 0.9944 & 0.9943 \\
        \bottomrule
    \end{tabular}
\end{table}

\paragraph{Invariance of $\overline{\mathrm{U}}$ Against Permutation of Samples}
In addition to the two requisites proposed above by \citep{widdowson2022resolving}, we introduce another crucial requirement: The model-level uniqueness score $\overline{\mathrm{U}}$ should be invariant against the permutation of generated samples. 
Ideally, the evaluation of a set of samples should remain independent of the order of generation; that is, whether structure $x$ is generated before structure $x'$ should be irrelevant to the final score. 
However, $\overline{\mathrm{U}}_\mathrm{smat}$ violates this fundamental principle. 
Consequently, varying $\overline{\mathrm{U}}_\mathrm{smat}$ scores can be derived from the identical set of crystals merely by shuffling their order.
To illustrate this phenomenon, consider a set of three generated samples $x$, $x'$, and $x''$, where $d_\mathrm{smat}(x, x') = d_\mathrm{smat}(x’, x'') = 0$, but a cumulative difference leads to $d_\mathrm{smat}(x, x'') = 1$.
If the generation order is $x \to x' \to x''$, $\overline{\mathrm{U}}_\mathrm{smat}$ equals to $\frac{1}{3}$, as only $x$ is deemed unique. 
However, if they are sampled as $x \to x'' \to x'$, both $x$ and $x''$ are regarded as unique, giving $\overline{\mathrm{U}}_\mathrm{smat} = \frac{2}{3}$.
Table~\ref{tab:invariance} shows that this is not just a theoretical concern. 
Using different seeds for shuffling yields different $\overline{\mathrm{U}}_\mathrm{smat}$ values for the 10k samples from CDVAE \citep{xie2022crystal}.
This lack of permutation invariance can be attributed to the fact that $d_\mathrm{smat}$ is not a true pseudometric. 
Specifically, it violates the triangle inequality. 
In contrast, none of the other crystal distance functions introduced in this paper has this flaw.
For a more rigorous discussion, please refer to \ref*{appendix:sec:sec:invariance} of the Supporting Information.

\subsection{Evaluation of Generative Models with Uniqueness and Novelty}
\label{sec:sec:un_experiment}

\paragraph{Experimental Settings}
To systematically investigate the influence of distance function selection on uniqueness and novelty assessments, we utilized samples generated by seven different models: CDVAE \citep{xie2022crystal}, DiffCSP \citep{jiao2023crystal}, DiffCSP++ \citep{jiao2024space}, MatterGen \citep{MatterGen2025}, Chemeleon-DNG \citep{park2025exploration}, ADiT \citep{joshi2025allatom}, and Chemeleon2 \citep{park2025guiding}, all of which were trained on the MP20 dataset. 
For each model, we generated 10,000 structures via de novo sampling of the crystal space.
Subsequently, uniqueness and novelty metrics were computed using the various distance functions.
Novelty was assessed relative to the MP20 training data.
It is important to emphasize that the primary objective of this section is to evaluate how the choice of distance function influences the uniqueness and novelty assessments. 
Therefore, we will not discuss why one model outperforms another in a specific metric in detail.

\begin{figure}
    \centering
    \includegraphics[width=\linewidth]{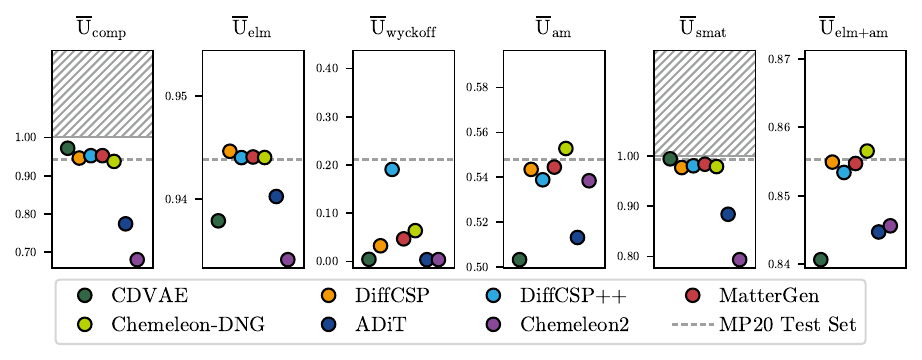}
    \caption{Uniqueness scores $\overline{\mathrm{U}}$ for each generative model based on different distances. The evaluation is based on 10k unconditioned samples. Each subplot shows $\overline{\mathrm{U}}$ computed with a specific distance. The horizontal gray dashed line represents $\overline{\mathrm{U}}$ of the MP20 test set. Since $\overline{\mathrm{U}}$ is guaranteed to be $\le$ 1.0, the area with $\overline{\mathrm{U}}>1$ is shaded in each subplot.}
    \label{fig:uni}
\end{figure}

\begin{figure}
    \centering
    \includegraphics[width=\linewidth]{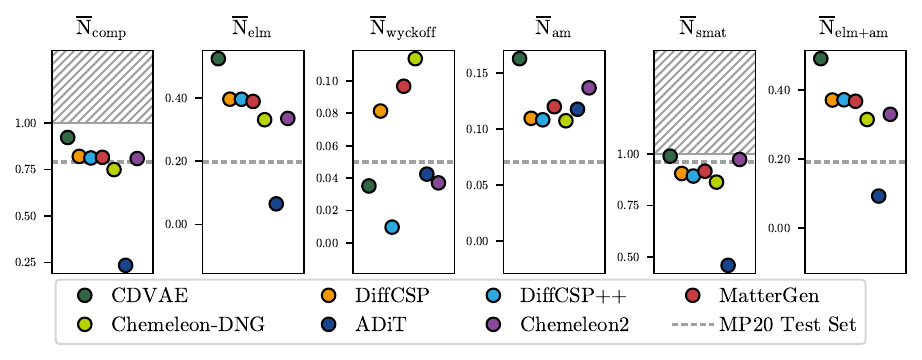}
    \caption{Novelty scores $\overline{\mathrm{N}}$ for each generative model based on different distances. Each subplot shows $\overline{\mathrm{N}}$ computed with a specific distance. The horizontal gray dashed line represents $\overline{\mathrm{N}}$ of the MP20 test set. Since $\overline{\mathrm{N}}$ is guaranteed to be $\le$ 1.0, the area with $\overline{\mathrm{N}}>1$ is shaded in each subplot.}
    \label{fig:nov}
\end{figure}

\begin{figure}
\captionsetup[subfigure]{
  justification=raggedright,
  singlelinecheck=false,
  format=hang,
}
    \centering
    \begin{minipage}{0.19\columnwidth}
        \includegraphics[width=0.95\columnwidth]{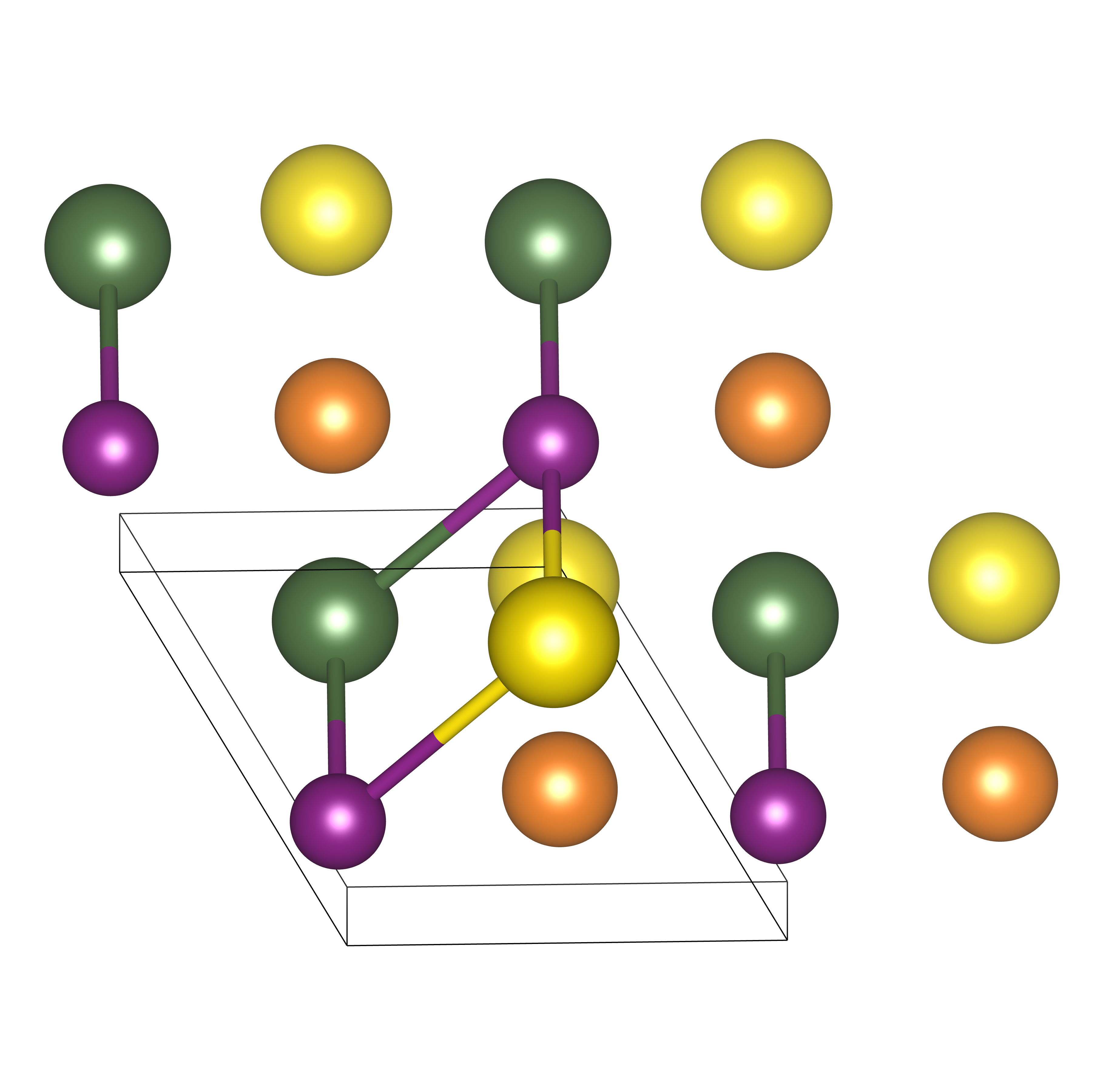}
        \subcaption{{\ColorUl{252,225,5}{\ce{Pr}}}{\ColorUl{73,114,58}{\ce{Er}}}{\ColorUl{251,123,21}{\ce{Mg}}}{\ColorUl{142,31,138}{\ce{I}}} \\[2pt] Cm \\ $E_\mathrm{hull} = 0.5338$ \\ $\mathrm{cN} = 0.7757$}
    \end{minipage}
    \begin{minipage}{0.19\columnwidth}
        \includegraphics[width=0.95\columnwidth]{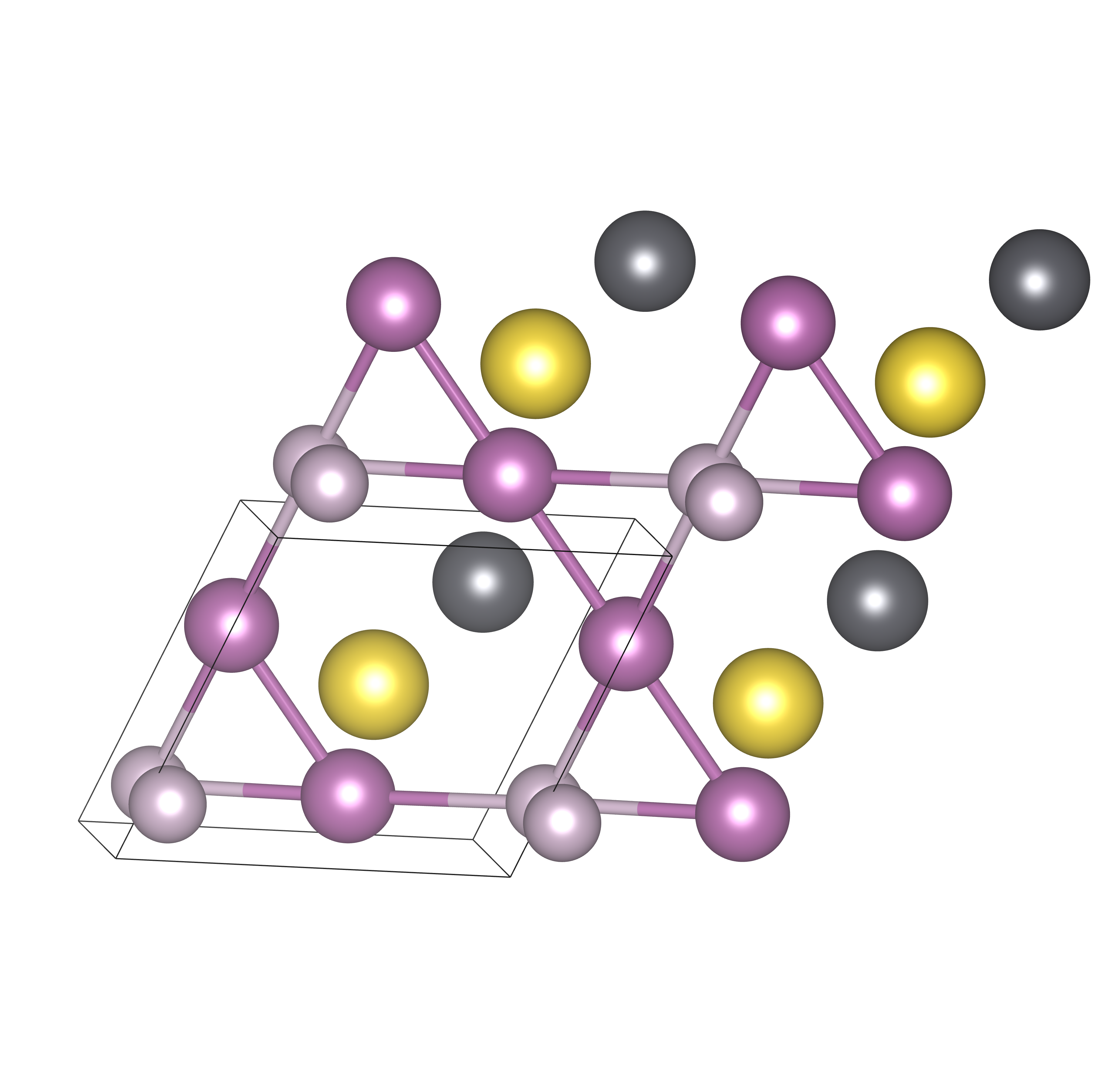}
        \subcaption{{\ColorUl{249,220,60}{\ce{Na}}}{\ColorUl{181,99,171}{\ce{Sc2}}}{\ColorUl{205,175,202}{\ce{Tc2}}}{\ColorUl{82,83,91}{\ce{Pb}}} \\[2pt] P1 \\ $E_\mathrm{hull} = 0.8257$ \\ $\mathrm{cN} = 0.7711$}
    \end{minipage}
    \begin{minipage}{0.19\columnwidth}
        \includegraphics[width=0.95\columnwidth]{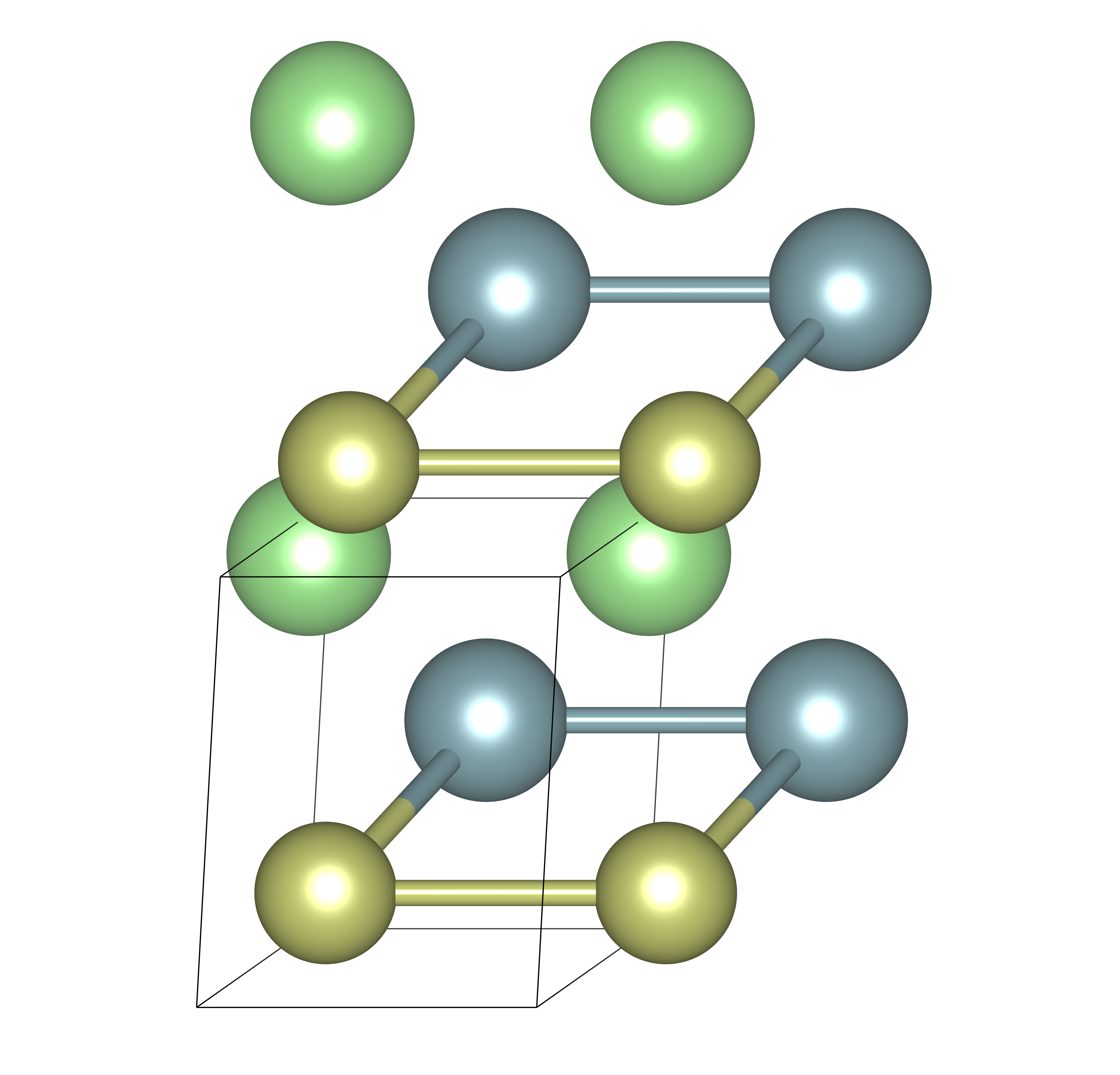}
        \subcaption{{\ColorUl{134,224,116}{\ce{Li}}}{\ColorUl{121,161,170}{\ce{U}}}{\ColorUl{201,206,114}{\ce{Ir}}} \\[2pt] P1 \\ $E_\mathrm{hull} = 0.6765$ \\ $\mathrm{cN} = 0.7689$}
    \end{minipage}
    \begin{minipage}{0.19\columnwidth}
        \includegraphics[width=0.95\columnwidth]{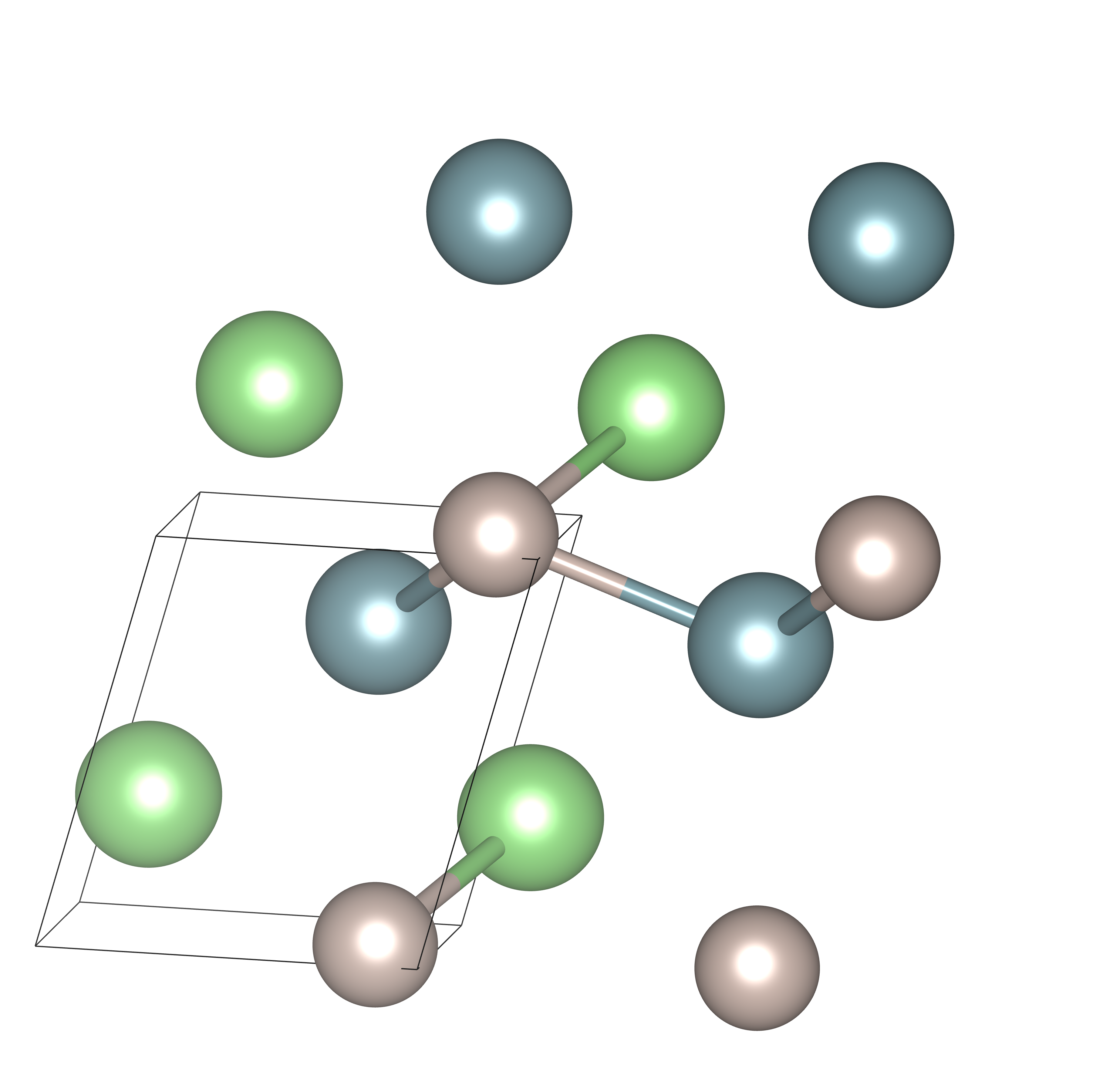}
        \subcaption{{\ColorUl{134,224,116}{\ce{Li}}}{\ColorUl{121,161,170}{\ce{U}}}{\ColorUl{207,183,173}{\ce{Ru}}} \\[2pt] P1 \\ $E_\mathrm{hull} = 0.6657$ \\ $\mathrm{cN} = 0.7682$}
    \end{minipage}
    \begin{minipage}{0.19\columnwidth}
        \includegraphics[width=0.95\columnwidth]{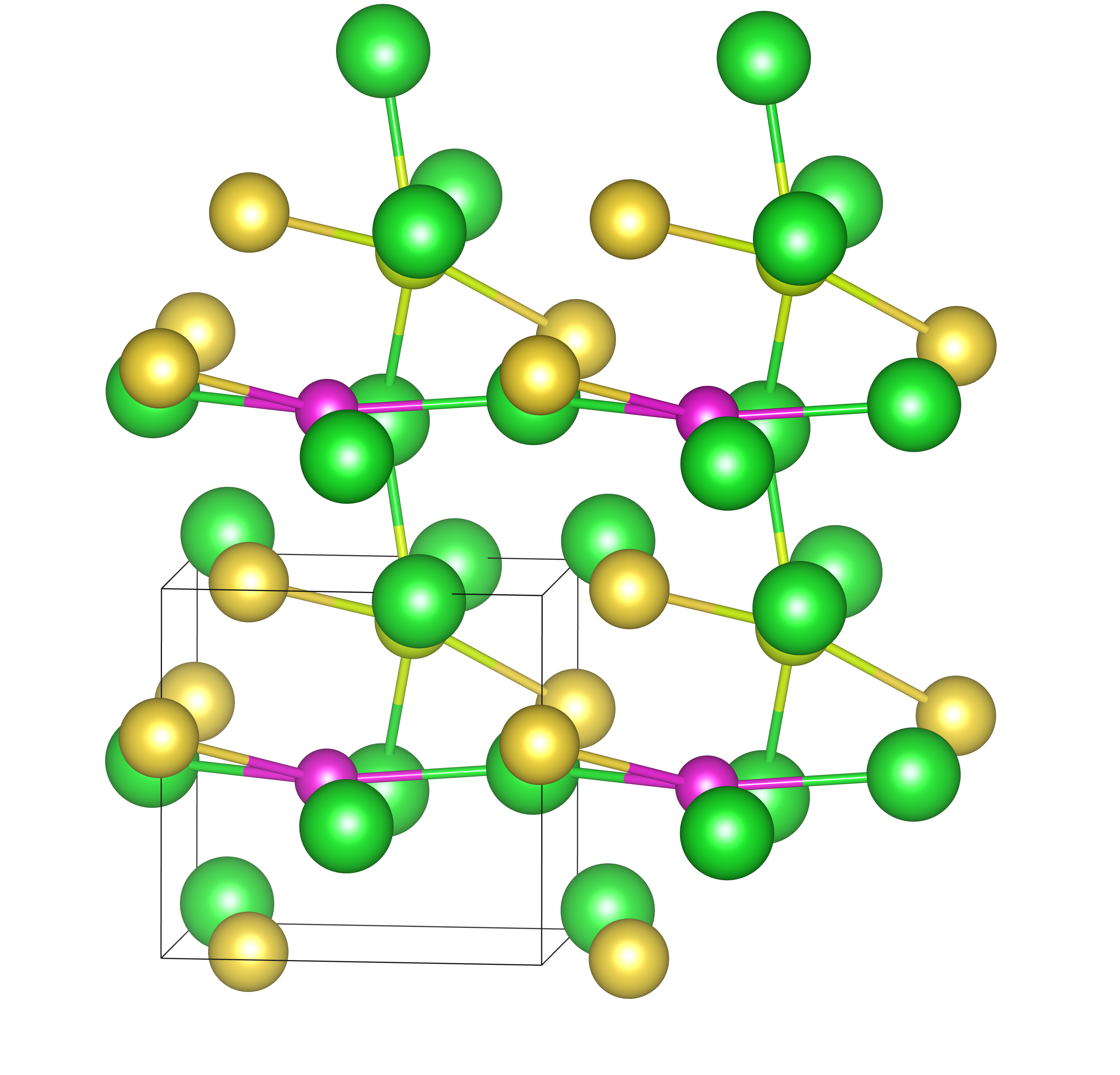}
        \subcaption{{\ColorUl{30,239,44}{\ce{Ba4}}}{\ColorUl{249,220,60}{\ce{Na2}}}{\ColorUl{209,252,6}{\ce{Ce}}}{\ColorUl{242,30,220}{\ce{Cd}}} \\[2pt] P1 \\ $E_\mathrm{hull} = 0.2781$ \\ $\mathrm{cN} = 0.7666$}
    \end{minipage}
    \caption{Top-5 most novel crystal samples from an early-days model, CDVAE, identified by $\mathrm{cN}_\mathrm{elm{+}am}$. Space groups are determined with the same tolerance threshold as the Materials Project. The energy above the convex hull [eV/atom] ($E_\mathrm{hull}$) values are computed against the Materials Project using the MACE-MPA-0 force field \citep{batatia2025foundation}. While CDVAE achieves the highest $\overline{\mathrm{N}}_\mathrm{elm{+}am}$ in Figure~\ref{fig:nov}, the high positive $E_\mathrm{hull}$ values suggest that this high novelty is due to the generation of physically implausible crystals.}
    \label{fig:nov_samples}
\end{figure}

\paragraph{Model-level Evaluation}
Figures~\ref{fig:uni} and \ref{fig:nov} show the model-level $\overline{\mathrm{U}}$ and $\overline{\mathrm{N}}$ for the various models, computed with different distances.
A key observation is the similarity in model distributions between the subplots for $\overline{\mathrm{U}}_\mathrm{comp}$ and $\overline{\mathrm{U}}_\mathrm{smat}$.
Closer inspection reveals that $\overline{\mathrm{U}}_\mathrm{comp}$ is quite high, with the difference $\overline{\mathrm{U}}_\mathrm{smat} - \overline{\mathrm{U}}_\mathrm{comp}$ remaining below 0.12 for all models.
This indicates that a substantial proportion of the generated samples have unique compositions, and that the widely adopted $d_\mathrm{smat}$ primarily serves as a compositional match checker, without examining the structures (Remember that $d_\mathrm{smat}$ first checks compositional identity). 
This observation holds true for $\overline{\mathrm{N}}_\mathrm{comp}$ and $\overline{\mathrm{N}}_\mathrm{smat}$ as well.
Another significant observation is that the continuous metrics uncover limitations in models that are not captured by conventional discrete metrics.
For instance, while CDVAE achieves the highest scores in $\overline{\mathrm{U}}_\mathrm{comp}$ and $\overline{\mathrm{U}}_\mathrm{smat}$, it exhibits poor performance in $\overline{\mathrm{U}}_\mathrm{elm}$ and $\overline{\mathrm{U}}_\mathrm{elm{+}am}$.
This suggests that, although CDVAE's output contains few exact duplicates in terms of discrete $d_\mathrm{comp}$ or $d_\mathrm{smat}$, the overall distribution in the continuous metric spaces is highly concentrated rather than dispersed, which hinders the generation of diverse candidates.
$\overline{\mathrm{U}}_\mathrm{comp}$ and $\overline{\mathrm{U}}_\mathrm{smat}$ tend to overestimate the diversity of such non-scattered samples, while $\overline{\mathrm{U}}_\mathrm{elm}$ and $\overline{\mathrm{U}}_\mathrm{elm{+}am}$ correctly penalize them.
A similar trend appears in structural uniqueness: DiffCSP++ ranks first in $\overline{\mathrm{U}}_\mathrm{wyckoff}$, but falls to fourth in $\overline{\mathrm{U}}_\mathrm{am}$. 
The model can generate diverse Wyckoff templates because it first samples a space group and then generates structures conditioned on the symmetry. 
Its poor performance in $\overline{\mathrm{U}}_\mathrm{am}$ indicates that diversity in Wyckoff templates does not translate directly to structural diversity defined by interatomic distances.
Likewise, the discrepancy between $\overline{\mathrm{N}}_\mathrm{wyckoff}$ and $\overline{\mathrm{N}}_\mathrm{am}$ rankings confirms that novelty in symmetry templates differs from novelty in terms of average interatomic distance.
Finally, a comparison against the MP20 test set baseline sheds light on the general trend of current generative models' capabilities.
While the majority of models surpass the test set in $\overline{\mathrm{N}}$, fewer models attain a higher $\overline{\mathrm{U}}$.
In other words, current models are effective at generating novel samples, yet face challenges in producing output as diverse as the test data (Note that stability is not considered here.) 
Generating diverse structures seems especially challenging since only Chemeleon-DNG exceeds the performance of the test set in $\overline{\mathrm{U}}_\mathrm{am}$.
The analysis presented here is limited to aggregated scores averaged over the samples generated by each model. \ref*{appendix:sec:sec:dist_cu_cn_elm+am} provides a finer-grained view by analyzing the pre-aggregation distributions of cU and cN for each model.

\paragraph{Sample-level Evaluation}
One practical advantage of continuous metrics is their ability to provide a granular ranking of candidates, rather than simply separating them into two groups of pass or fail.
Although their primary purpose is to weight each sample’s contribution to the final model score according to its quality, they can also be used as soft screening scores to identify samples for further analysis.
Figure~\ref{fig:nov_samples} shows the most novel samples identified by $\mathrm{cN}_\mathrm{elm{+}am}$ within the output from CDVAE, a model which scored the highest average novelty $\overline{\mathrm{N}}_\mathrm{elm{+}am}$.
The identified samples do not have a typical composition or structure, confirming that $\mathrm{cN}_\mathrm{elm{+}am}$ can successfully measure the dissimilarity between the generated and training data samples.
However, closer inspection reveals that the identified crystals are also physically unrealistic.
Crystal (a) fails to pass the SMACT chemical validity filter, which considers the charge neutrality and the Pauling electronegativity requirements of the compound \citep{davies2016computational}. 
While crystals (b) through (d) are SMACT-valid, they are predicted to be thermodynamically unstable.
Only Crystal (e) possesses a SMACT-valid composition and a relatively moderate $E_\mathrm{hull}$.
These findings suggest that CDVAE attains its high $\overline{\mathrm{N}}_\mathrm{elm{+}am}$ score largely by generating physically implausible crystals. 
This result underscores the fact that uniqueness and novelty alone are insufficient for a comprehensive evaluation.
It is imperative to assess the feasibility of samples simultaneously, an issue we address in the subsequent section.

\section{Stability and SUN}
As observed in Section~\ref{sec:sec:un_experiment}, uniqueness and novelty alone are insufficient to evaluate generative models. 
It is possible for models to readily achieve high scores in these metrics merely by generating diverse yet physically implausible crystals.
Consequently, a third criterion is required to assess the plausibility of candidate crystals. 
For this purpose, thermodynamic stability is widely employed as a computable proxy.
Section~\ref{sec:sec:s_definition} discusses the limitations of the conventional binary stability score and elucidates how our proposed continuous score remedies these issues. 
The proposed continuous score is designed to spread scores across generated samples for benchmarking. 
It should not be interpreted as a quantitative probability of synthesis or as a replacement for higher-fidelity DFT and experimental validation.
Subsequently, Section~\ref{sec:sec:sun_definition} combines all of our continuous metrics into a single cSUN metric, offering a continuous alternative to the widely used SUN metric. 
Finally, Section~\ref{sec:sec:sun_experiment} presents an evaluation of the generative models utilizing our proposed cSUN metric.

\subsection{Continuous Stability}
\label{sec:sec:s_definition}

\begin{figure}
    \centering
    \includegraphics[width=\linewidth]{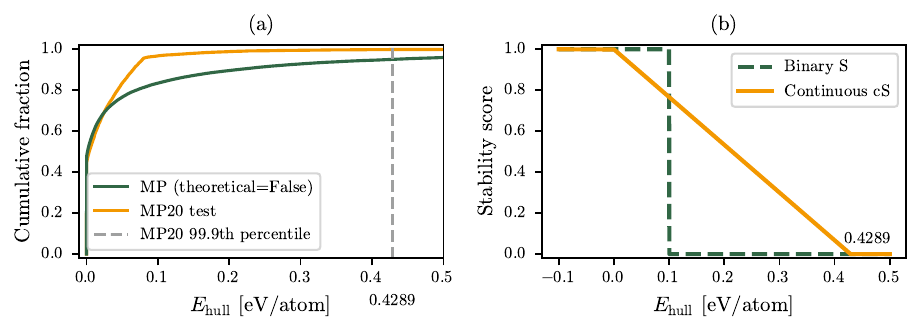}
    \caption{(a): Cumulative distribution of $E_\mathrm{hull}$. Approximately 20\% of the experimentally synthesized crystals in the Materials Project (theoretical=False) have $E_\mathrm{hull}$ values greater than the typical threshold of 0.1 [eV/atom].  (b): Comparison of stability score functions. While S drops down at $E_\mathrm{hull} = 0.1$ [eV/atom], cS decreases gradually from $E_\mathrm{hull} = 0$ [eV/atom] to $E_\mathrm{hull} = 0.4289$ [eV/atom]. The value of 0.4289 is the 99.9th percentile in the MP20 test data shown in (a).}
    \label{fig:ehull_stability}
\end{figure}

\begin{figure}
    \centering
    \includegraphics[width=\linewidth]{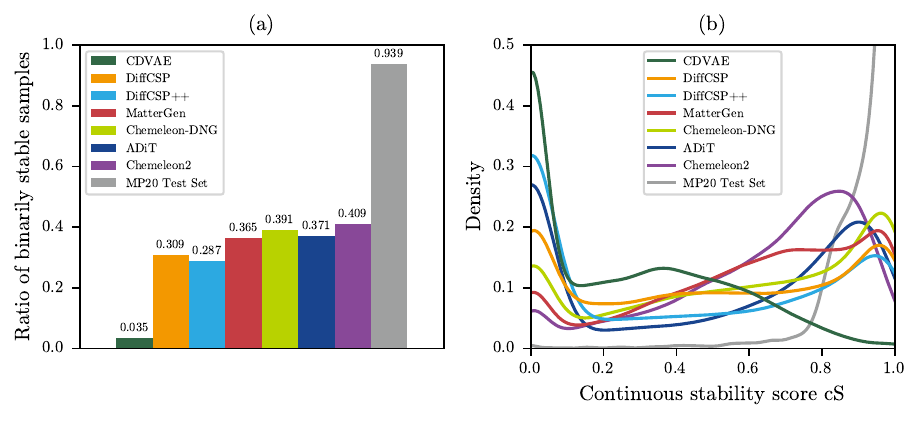}
    \caption{(a): Ratio of samples whose binary stability score $\mathrm{S}(x)$ is one. The majority of the generated samples are classified as unstable (S=0), which means they are simply discarded during the evaluation process. (b): Distribution of continuous stability score $\mathrm{cS}(x)$. cS offers a smoother score distribution, and fewer samples are merely removed.}
    \label{fig:stability_b_c}
\end{figure}

Most previous studies have evaluated the stability of the sample $x_i$ using the following binary criterion: 
\begin{equation}
    \label{eq:binary-s}
    \mathrm{S}(x_i) \coloneqq \begin{cases}
        1 & \text{if $E_\mathrm{hull}(x_i) \le \tau$} \\
        0 & \text{otherwise},
    \end{cases}
\end{equation}
where $E_\mathrm{hull}(x_i)$ is the energy above the convex hull, computed relative to a set of known crystals (e.g., the Materials Project).
The threshold $\tau$ is typically set to around $0.1$ [eV/atom].
This value can be determined through several approaches, such as a statistical analysis of previously synthesized crystals \citep{sun2016thermodynamic}, the limits of DFT accuracy \citep{gong2022calibrating}, or theoretical estimates of entropy contributions to Gibbs free energy at finite temperatures \citep{aykol2018thermodynamic}.
However, a major limitation of this binary approach is its failure to capture the gradation of stability.
For example, $E_\mathrm{hull}=0.11$ [eV/atom] and $E_\mathrm{hull}=1.0$ [eV/atom] yield the same score, as do $E_\mathrm{hull}=0.0$ [eV/atom] and $E_\mathrm{hull}=0.09$ [eV/atom].
Furthermore, assigning a zero score to samples slightly above $\tau$ risks discarding novel candidates that hold significant potential for materials discovery.
While excluding thermodynamically implausible structures is necessary, an overly stringent threshold may inadvertently filter out all but the most conservative (low-novelty) samples.
Given that $E_\mathrm{hull}$ is just a computable proxy, samples slightly above $\tau$ may still be accessible, and they are expected to be more novel.
In fact, a non-negligible fraction of experimentally synthesized crystals in the Materials Project possess $E_\mathrm{hull}$ values exceeding 0.1 [eV/atom] (Figure~\ref{fig:ehull_stability}(a); see also \citep{sun2016thermodynamic, aykol2018thermodynamic}).
To address these issues, we introduce a continuous stability (cS):
\begin{equation}
    \label{eq:continuous-s}
    \mathrm{cS}(x_i) \coloneqq \begin{cases}
        1 & \text{if $E_\mathrm{hull}(x_i) \le 0$} \\
        1 - \frac{E_\mathrm{hull}(x_i)}{\tau} & \text{if $0 < E_\mathrm{hull}(x_i) \le \tau$} \\
        0 & \text{otherwise}.
    \end{cases}
\end{equation}
Figure~\ref{fig:ehull_stability}(b) provides a visual comparison of S and cS. 
Unlike the binary S, cS continuously decreases as $E_\mathrm{hull}$ increases. 
The threshold $\tau$ is a hyperparameter.
In this study, it is set to 0.4289 [eV/atom], which is the 99.9th percentile of the $E_\mathrm{hull}$ distribution of the MP20 test set (Figure~\ref{fig:ehull_stability}(a)).
Since the original MP20 data is outdated, we recalculated their $E_\mathrm{hull}$ values using the latest Materials Project database.
cS should be regarded as a soft screening score, not as a quantitative measure of synthesizability, because it uses a relatively permissive threshold of $\tau = 0.4289$ [eV/atom]. 
This loose threshold reduces the number of zero-score samples by assigning non-zero cS values to a broader range of generated crystals. 
It allows crystals with non-negligible $E_\mathrm{hull}$ to contribute to the evaluation, consistent with the observation that approximately $4.9\%$ of experimentally synthesized Materials Project entries satisfy $E_\mathrm{hull} \ge 0.4289$ [eV/atom]. 
However, we acknowledge that a permissive choice of $\tau$ may also increase the false-positive rate; that is, inaccessible crystals may still be assigned non-zero scores. 
We therefore provide a detailed sensitivity analysis of $\tau$ in \ref*{appendix:sec:sec:cs_sensitivity}. 
We also examine alternative functional forms of cS.
Figure~\ref{fig:stability_b_c} illustrates the distributions of S and cS across different generative models. 
When employing the binary S, a substantial proportion (60 to 70\%) of the generated samples is simply discarded (96.5\% for CDVAE). 
In contrast, the continuous cS offers a smoother score distribution.

\subsection{Continuous SUN metric}
\label{sec:sec:sun_definition}

\begin{figure}
    \centering
    \includegraphics[width=\linewidth]{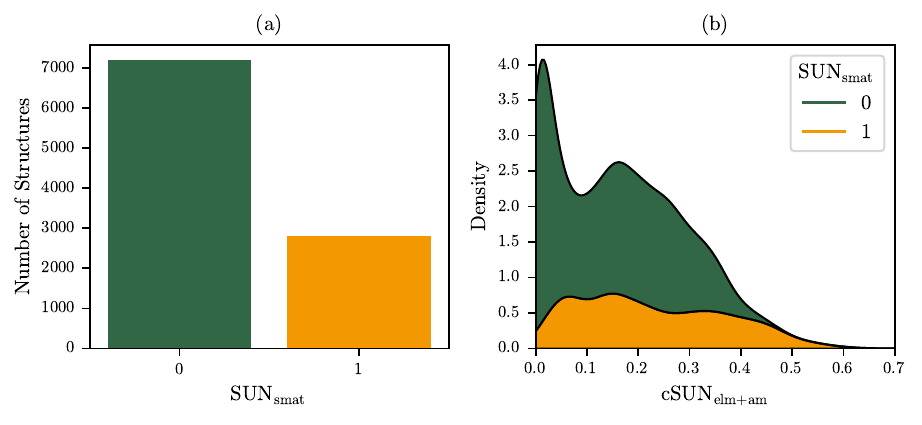}
    \caption{(a): Comparison of the $\mathrm{SUN}_\mathrm{smat}$ score for 10k crystal samples from MatterGen. (b): Distribution of the $\mathrm{cSUN}_\mathrm{elm{+}am}$ score for 10k samples from MatterGen. $\mathrm{SUN}_\mathrm{smat}$ and $\mathrm{cSUN}_\mathrm{elm{+}am}$ positively correlate with each other, but $\mathrm{cSUN}_\mathrm{elm{+}am}$ offers a more detailed ranking of candidates.}
    \label{fig:sun_csun_mattergen}
\end{figure}

While we have defined uniqueness, novelty, and stability separately so far, it is also helpful to have a single unified metric to compare models.
The most widely used metric, SUN, is defined as the product of binary S, U, and N introduced in Equations~\eqref{eq:sample-level-u}, \eqref{eq:sample-level-n}, and \eqref{eq:binary-s}:
\begin{equation*}
    \mathrm{SUN}(x_i) \coloneqq \mathrm{S}(x_i) \cdot \mathrm{U}(x_i) \cdot \mathrm{N}(x_i) \in \{0, 1\}.
\end{equation*}
For U and N, the StructureMatcher distance $d_\mathrm{smat}$ is typically used \citep{MatterGen2025, miller2024flowmm, levy2025symmcd, kazeev2025wyckoff, sriram2024flowllm, de2026generative, joshi2025allatom, cornet2025kinetic, hollmer2025open, park2025guiding, xu2025plaid++, antunes2024crystal, yan2024invariant, gan2025matllmsearchcrystalstructurediscovery}.
The model-level SUN score is calculated as the average of sample-level scores, i.e. $\overline{\mathrm{SUN}} \coloneqq \sum_{i=1}^{n} \mathrm{SUN}(x_i)$.
Given that the SUN metric is derived from the binary S, U, and N, it inevitably inherits all their limitations discussed in Section~\ref{sec:un} and \ref{sec:sec:s_definition}. 
In addition, it only separates the candidate samples into two rough classes: SUN=1 and SUN=0 (Figure~\ref{fig:sun_csun_mattergen}(a)). 
Consequently, candidates with superior stability, uniqueness, and novelty contribute to the final model score in the same way as marginal candidates that lie just above the threshold.
This binary nature also presents challenges when selecting the most promising samples for further investigation: Of the $\sim$3000 samples with SUN = 1 in Figure~\ref{fig:sun_csun_mattergen}(a), which are the most promising?
To address these shortcomings, we propose the continuous SUN (cSUN), defined as the product of cS, cU, and cN introduced in Equations \eqref{eq:sample-level-u}, \eqref{eq:sample-level-n}, and \eqref{eq:continuous-s}:
\begin{equation*}
    \mathrm{cSUN}(x_i) \coloneqq \mathrm{cS}(x_i)^{w_S} \cdot \mathrm{cU}(x_i)^{w_U} \cdot \mathrm{cN}(x_i)^{w_N} \in [0, 1].
\end{equation*}
Here, $w_S$, $w_U$, and $w_N$ are hyperparameters that enable the prioritization of specific components over others.
For instance, assigning a high value to $w_S$ favors stable crystals.
In this study, all weights are set to 1 unless otherwise stated.
It is important to note that such weighting schemes are not applicable to the binary SUN metric, because $0^x$ and $1^x$ are (almost) invariant to the value of $x$.
Consistent with the binary SUN, the model-level cSUN score $\overline{\mathrm{SUN}}$ is calculated as the average of the sample-level cSUN scores.
Hereafter, we adopt the notation $\mathrm{SUN}_\mathrm{name}$ (or $\mathrm{cSUN}_\mathrm{name}$) and $\overline{\mathrm{SUN}}_\mathrm{name}$ to denote metrics based on $d_\mathrm{name}$.
Figure~\ref{fig:sun_csun_mattergen}(b) shows the $\mathrm{cSUN}_\mathrm{elm{+}am}$ score distribution for 10k samples generated by MatterGen.
A positive correlation is observed between $\mathrm{SUN}_\mathrm{smat}$ and $\mathrm{cSUN}_\mathrm{elm{+}am}$, indicating that crystals with a $\mathrm{SUN}_\mathrm{smat}$ value of 1 tend to exhibit higher $\mathrm{cSUN}_\mathrm{elm{+}am}$ scores.
In fact, the probability of superiority (PS) is approximately 0.67, where PS is defined as:
\begin{equation*}
     \mathrm{PS} \coloneqq p\left(\mathrm{cSUN}_\mathrm{elm{+}am}(x)\big\rvert_{\mathrm{SUN}_\mathrm{smat}(x)=1} > \mathrm{cSUN}_\mathrm{elm{+}am}(x)\big\rvert_{\mathrm{SUN}_\mathrm{smat}(x)=0}\right).
\end{equation*}
This indicates that a binary-SUN-positive sample has a higher $\mathrm{cSUN}_\mathrm{elm{+}am}$ value than a binary-SUN-negative sample with probability 0.67.
The cSUN metric successfully smooths out the histogram in Figure~\ref{fig:sun_csun_mattergen}(a), offering a detailed ranking of the samples.

\subsection{Reference Distributions Used in the Proposed Metrics}
As the proposed metrics compare generated samples against different sets of crystals, we briefly summarize the role of each reference distribution before presenting the experimental results. 
For uniqueness, the relevant reference set is the generated set itself, within which pairwise distances are computed. 
Binary uniqueness is sensitive to the number of generated samples, as exact matches become more likely in larger sample sets. 
Continuous uniqueness is expected to be less sample-size dependent, because it measures the average distance from each sample to the remaining generated samples. 
For novelty, the reference distribution is the training set; here, we use the MP20 training set, comprising 27,136 metastable crystals with up to 20 atoms in the primitive unit cell \cite{xie2022crystal}. 
For stability, the reference is the convex hull constructed from the Materials Project database, where a hull built from more structures should, in principle, yield a more accurate stability assessment. 
The MP20 test set is used as a baseline for model-level comparison and, in some analyses, to calibrate benchmark hyperparameters, including the default cS threshold $\tau$ and the compositional/structural mixing weights in $d_\mathrm{elm+am}$. 
One analysis not considered in the main study, but potentially of interest, is a comparison of sample distributions across different generative models. Preliminary results are provided in \ref*{appendix:sec:sec:model_dist}.

\subsection{Evaluation of Generative Models with SUN and cSUN}
\label{sec:sec:sun_experiment}

\paragraph{Experimental Settings}
Here, we evaluate generative models using SUN and cSUN. 
The sampling procedures and model family employed here are identical to those detailed in Section~\ref{sec:sec:un_experiment}.
Since it is computationally prohibitive to calculate the thermodynamic stability of thousands of samples with DFT, the MACE-MPA-0 force field was used for all energy calculations \citep{batatia2025foundation}.

\begin{figure}
    \centering
    \includegraphics[width=0.95\linewidth]{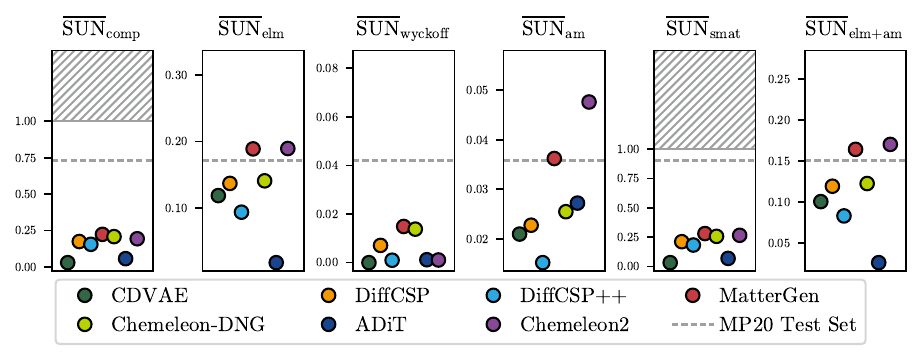}
    \caption{$\overline{\mathrm{SUN}}$ scores for each generative model based on different distances. The evaluation is based on 10k unconditioned samples. Each subplot shows $\overline{\mathrm{SUN}}$ computed with a specific distance. The horizontal gray dashed line represents $\overline{\mathrm{SUN}}$ for the MP20 test set. Since $\overline{\mathrm{SUN}}$ is guaranteed to be $\le$ 1.0, the area with $\overline{\mathrm{SUN}}>1$ is shaded in each subplot.}
    \label{fig:sun}
\end{figure}

\paragraph{Model-level Evaluation}
Figure~\ref{fig:sun} presents the model-level $\overline{\mathrm{SUN}}$ scores derived from different distance functions.
When evaluated with binary $\overline{\mathrm{SUN}}$ ($\overline{\mathrm{SUN}}_\mathrm{comp}$, $\overline{\mathrm{SUN}}_\mathrm{wyckoff}$, or $\overline{\mathrm{SUN}}_\mathrm{smat}$), the MP20 test set baseline significantly outperforms the generative models.
This is largely attributable to the fact that 94\% of the test set satisfies $E_\mathrm{hull} \le 0.1$ [eV/atom], a rate unmatched by the generative models (Figure~\ref{fig:stability_b_c}(a)).
In contrast, Chemeleon2 and MatterGen surpass the test set performance under continuous $\overline{\mathrm{SUN}}$ ($\overline{\mathrm{SUN}}_\mathrm{elm}$, $\overline{\mathrm{SUN}}_\mathrm{am}$, or $\overline{\mathrm{SUN}}_\mathrm{elm{+}am}$).
This is likely due to the significant contribution of generated samples that possess $E_\mathrm{hull}$ above 0.1 [eV/atom] but below $\tau =0.4289$ [eV/atom] and exhibit high uniqueness and novelty.
While such samples may include promising candidates for materials discovery, they make no contribution to the $\overline{\mathrm{SUN}}$ score when binary S and binary SUN are employed.
Note that this discussion is for the default configuration of cS. 
\ref*{appendix:sec:sec:cs_sensitivity} assesses the sensitivity of the results to the choice of the functional form and threshold value of cS. 
Overall, the model rankings on $\overline{\mathrm{SUN}}_\mathrm{elm+am}$ remain largely stable across different cS configurations, except for CDVAE, whose stability score distribution differs markedly from those of the other models (Figure~\ref{fig:stability_b_c}). 
Whether a model exceeds the MP20 test set performance depends on how strongly the cS function penalizes samples with non-negligible $E_\mathrm{hull}$.

\begin{figure}
    \centering
    \includegraphics[width=\linewidth]{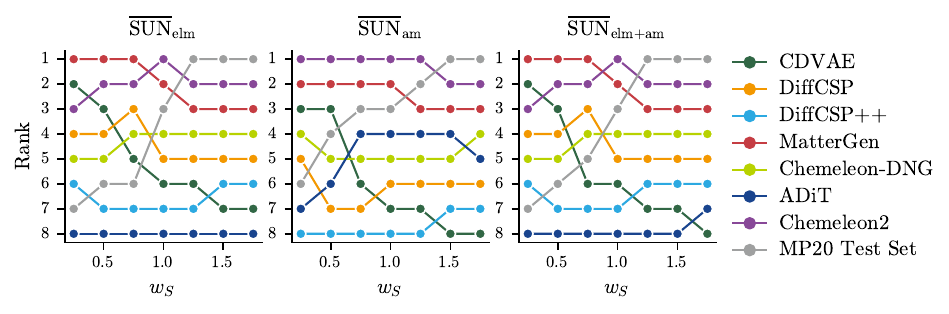}
    \caption{Ranking of models based on $\overline{\mathrm{SUN}}$ scores using different continuous distances with different weights for the underlying cSUN. Specifically, we set $(w_S, w_U, w_N) = (w_S, 1, 2-w_S)$. A larger $w_S$ indicates that stability is prioritized over novelty during evaluation.}
    \label{fig:sun_weight}
\end{figure}

One advantage of cSUN over SUN lies in the inclusion of tunable weight hyperparameters $(w_S, w_U, w_N)$.
This high flexibility allows users to customize the weights according to their practical needs.
Figure~\ref{fig:sun_weight} demonstrates how the relative ranking of models changes when these weights are adjusted. 
Specifically, we tested seven different weight values for stability and novelty, which usually present a trade-off. 
The results show that increasing $w_S$—that is, prioritizing stability—leads to an improvement in the rank of the MP20 test set.
This is because the MP20 test samples are generally more stable than those generated from ML models.
Conversely, the ranking of CDVAE declines as $w_S$ increases, indicating that CDVAE samples are the least stable.
Interestingly, MatterGen and Chemeleon2 consistently maintain top positions (excluding the test set baseline), demonstrating their effectiveness in generating samples that are both novel and stable.
\ref*{appendix:sec:sec:csun_weight_sensitivity} provides a more comprehensive analysis of how the component weights affect model rankings by varying all three weights simultaneously.
In brief, $w_S$ has the largest impact, and Chemeleon2 and MatterGen consistently rank highest among the generative models when stability is considered ($w_S > 0$).

\begin{figure}
\captionsetup[subfigure]{
  justification=raggedright,
  singlelinecheck=false,
  format=hang,
}
    \centering
    \begin{minipage}{0.19\columnwidth}
        \includegraphics[width=0.95\columnwidth]{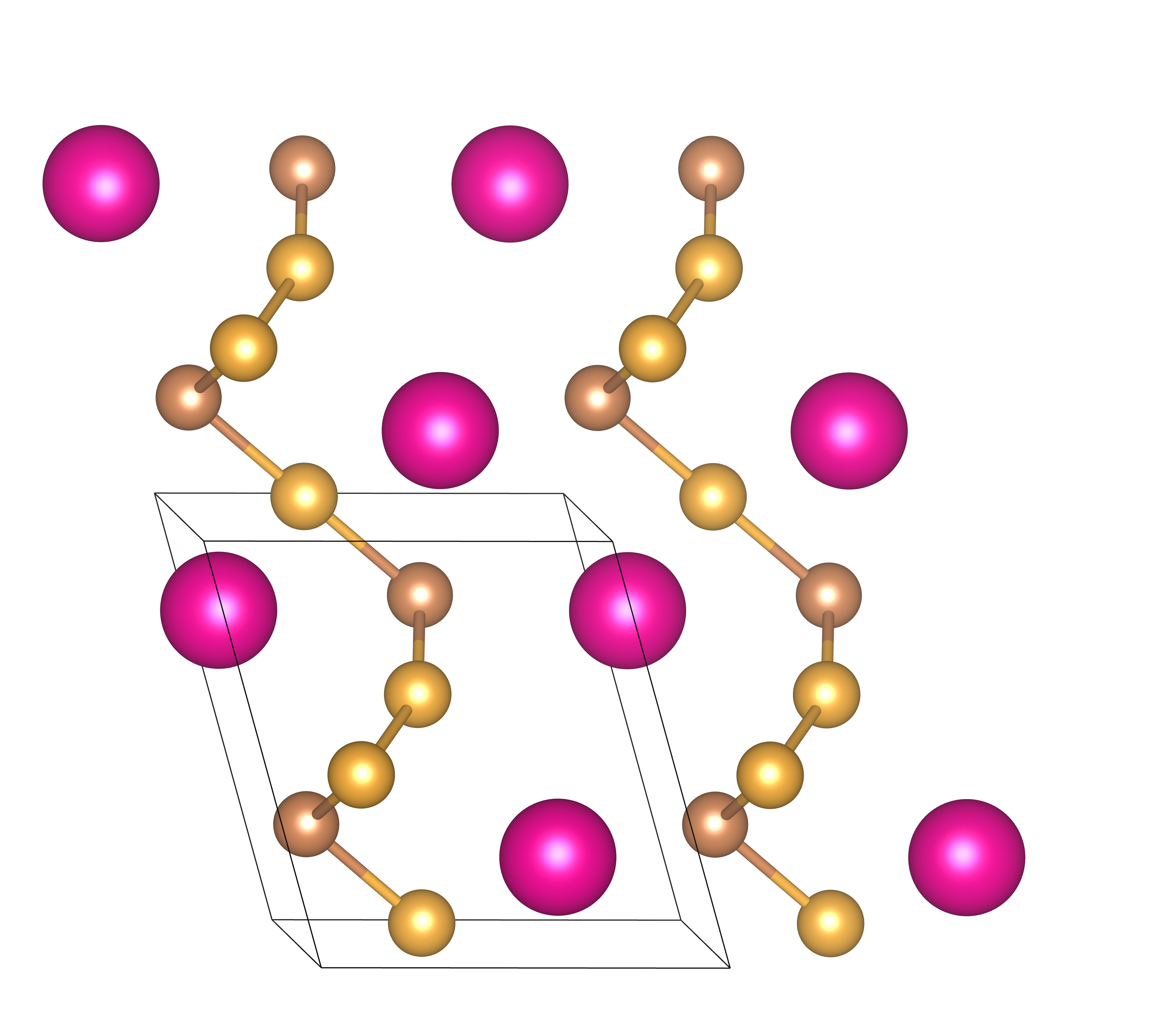}
        \subcaption{{\ColorUl{255,0,153}{\ce{Rb2}}}{\ColorUl{215,131,79}{\ce{Sb2}}}{\ColorUl{254,178,56}{\ce{Au3}}} \\[2pt] C2/m \\ $E_\mathrm{hull} = -0.0222$ \\ $\mathrm{cSUN} = 0.6325$}
    \end{minipage}
    \begin{minipage}{0.19\columnwidth}
        \includegraphics[width=0.95\columnwidth]{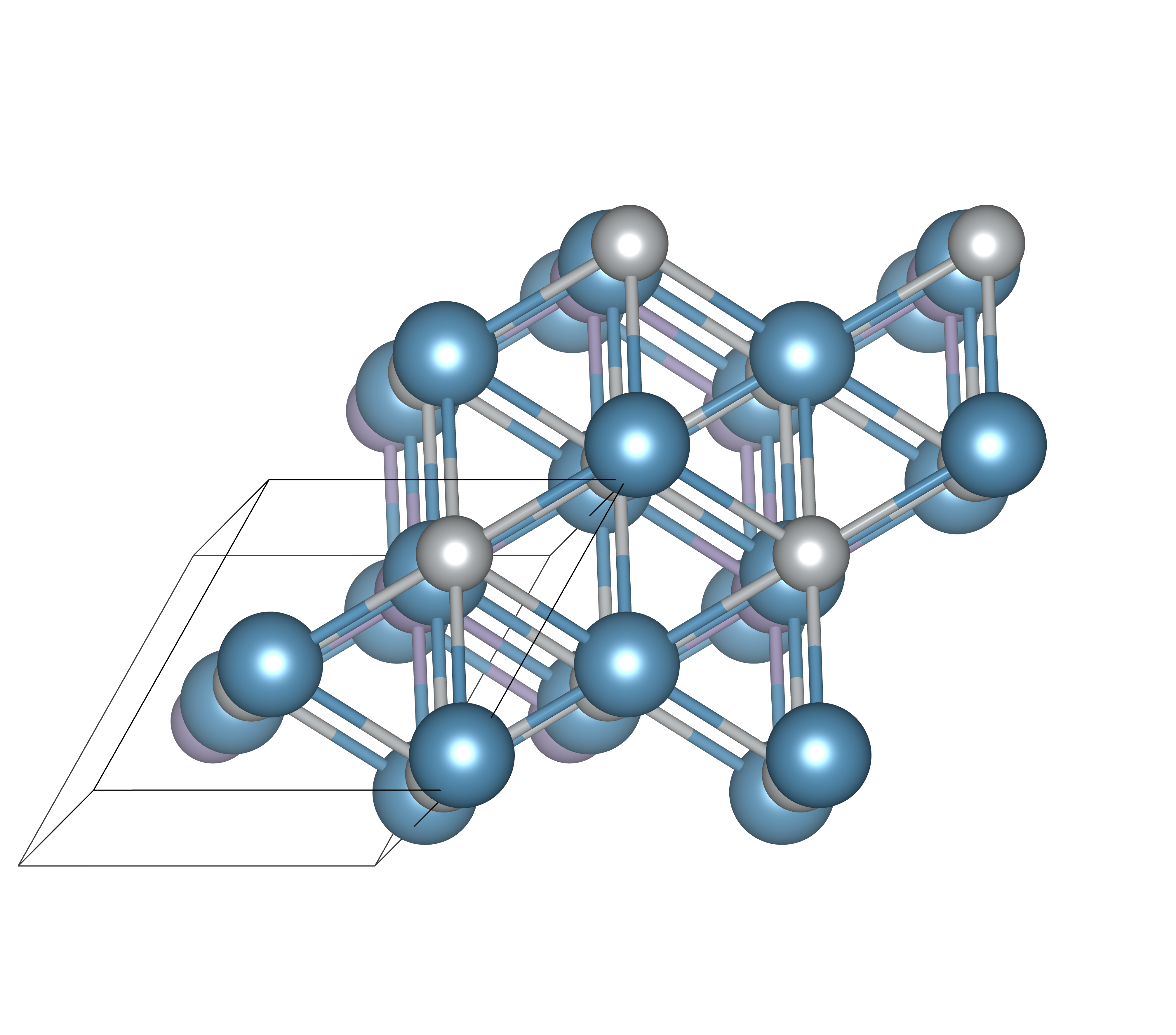}
        \subcaption{{\ColorUl{90,150,189}{\ce{Ca6}}}{\ColorUl{183,187,189}{\ce{Ag3}}}{\ColorUl{154,142,185}{\ce{Sn2}}} \\[2pt] P-3m1 \\ $E_\mathrm{hull} = 0.0274$ \\ $\mathrm{cSUN} = 0.5934$}
    \end{minipage}
    \begin{minipage}{0.19\columnwidth}
        \includegraphics[width=0.95\columnwidth]{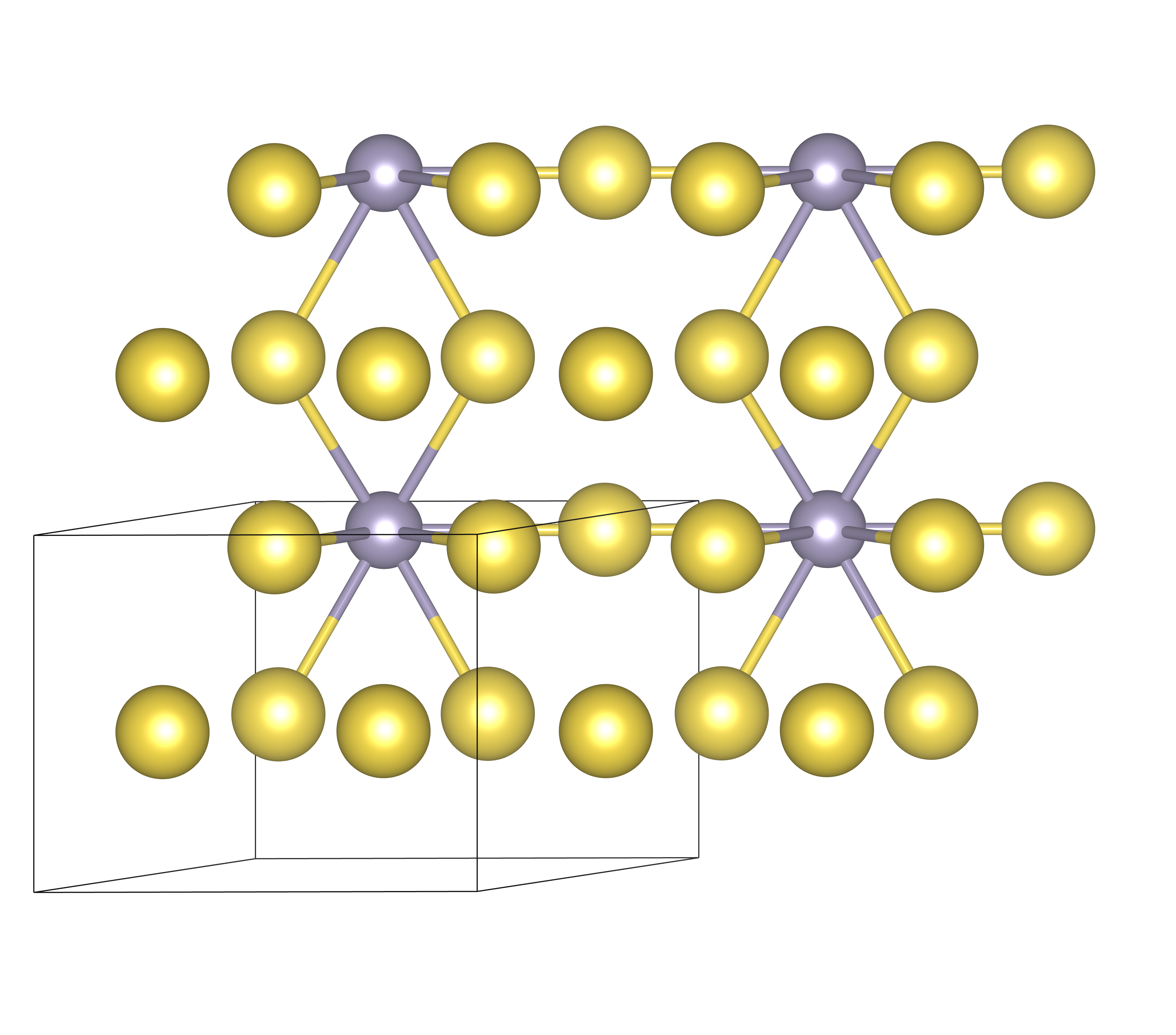}
        \subcaption{{\ColorUl{249,220,60}{\ce{Na7}}}{\ColorUl{154,142,185}{\ce{Sn}}} \\[2pt] P-6m2 \\ $E_\mathrm{hull} = 0.0503$ \\ $\mathrm{cSUN} = 0.5892$}
    \end{minipage}
    \begin{minipage}{0.19\columnwidth}
        \includegraphics[width=0.95\columnwidth]{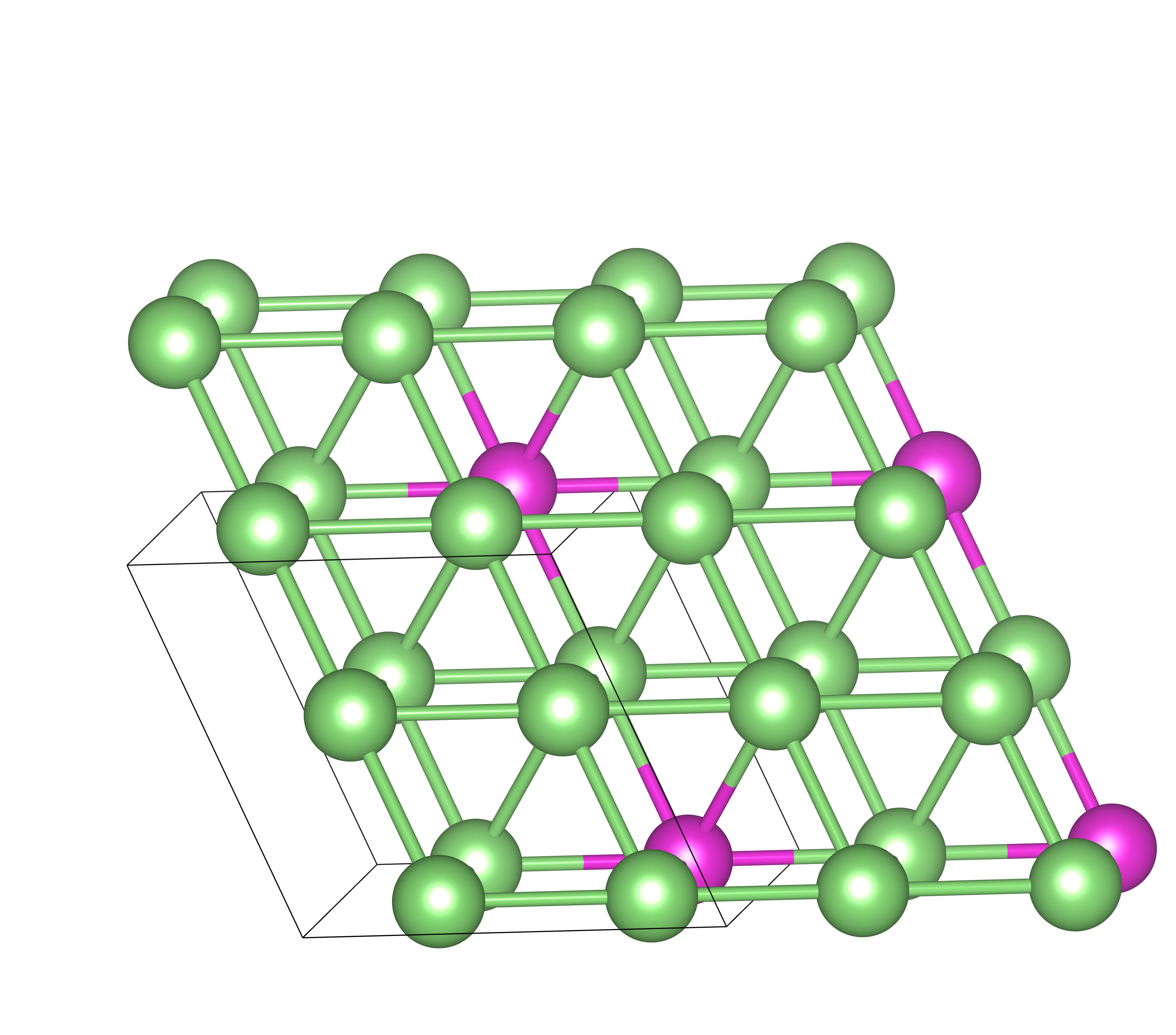}
        \subcaption{{\ColorUl{134,224,116}{\ce{Li7}}}{\ColorUl{242,30,220}{\ce{Cd}}} \\[2pt] Im-3m \\ $E_\mathrm{hull} = 0.0079$ \\ $\mathrm{cSUN} = 0.5882$}
    \end{minipage}
    \begin{minipage}{0.19\columnwidth}
        \includegraphics[width=0.95\columnwidth]{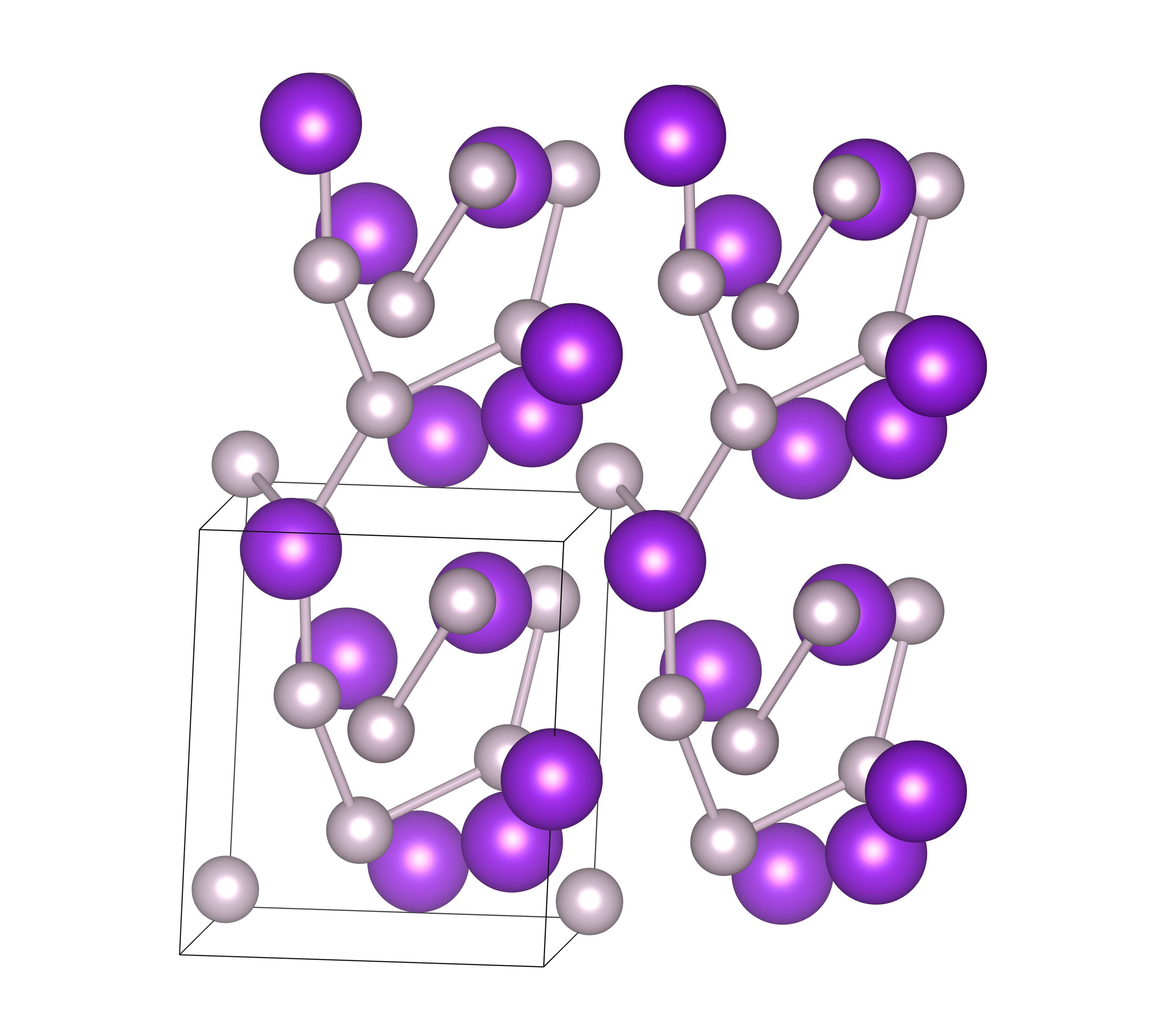}
        \subcaption{{\ColorUl{161,33,246}{\ce{K6}}}{\ColorUl{211,183,203}{\ce{Hg8}}} \\[2pt] P1 \\ $E_\mathrm{hull} = 0.0367$ \\ $\mathrm{cSUN} = 0.5854$}
    \end{minipage}
    \caption{Top-5 samples from MatterGen identified by $\mathrm{cSUN}_\mathrm{elm{+}am}$. Space groups are determined with the same tolerance threshold as the Materials Project. $E_\mathrm{hull}$ [eV/atom] values are computed against the Materials Project. Unlike Figure~\ref{fig:nov_samples}, these candidates exhibit significantly greater structural plausibility, since stability is considered.}
    \label{fig:vsun_samples}
\end{figure}

\paragraph{Sample-level Evaluation}
Another strength of cSUN is its ability to provide detailed sample rankings.
When used as a soft screening score, it helps identify promising samples for further inspection.
Figure~\ref{fig:vsun_samples} visualizes the SMACT-valid samples from MatterGen with the highest $\mathrm{cSUN}_\mathrm{elm{+}am}$ scores.
Compared to Figure~\ref{fig:nov_samples}, which does not consider stability, these crystals exhibit more realistic structures with lower $E_\mathrm{hull}$ values and higher symmetry.
Crystal (a) appears to be a Zintl phase where \ce{Au} and \ce{Sb} share electrons supplied by \ce{Rb}. 
Crystal (b) is an intermetallic compound with \ce{Sn} vacancies. 
Crystals (c) and (d) are also intermetallic compounds where \ce{Sn} and \ce{Cd} are surrounded by neatly aligned \ce{Na} and \ce{Li}, respectively. 
Finally, crystal (e) seems to be an amalgam in which \ce{K+} ions are surrounded by \ce{Hg} chains built with electrons supplied by \ce{K}.
Notably, all five crystals possess novel compositions that are not included in the MP20 training set.
While crystals (b) through (d) might initially appear to be variations of known structures with trivial defects, our careful inspection revealed that their defect-free counterparts (\ce{Ca2Ag1Sn1} without a Sn vacancy, a hexagonal Na, and a cubic Li) are not included in the MP20 training set.
Had these counterparts been present, the crystals would have yielded a lower $\mathrm{cN}_\mathrm{elm{+}am}$ score, and thus a lower $\mathrm{cSUN}_\mathrm{elm{+}am}$ score.
It is also worth noting that the identified crystals are not biased toward a particular composition or structure. 
These results demonstrate that our $\mathrm{cSUN}_\mathrm{elm{+}am}$ metric is effective in identifying a diverse range of stable and novel samples.

\section{Reinforcement Learning with SUN and cSUN}

Thus far, the discussion has focused on the improvement of evaluation metrics for generative models. 
Since these metrics essentially codify the desired characteristics of crystals, models attaining higher scores are deemed more useful. 
However, a potential limitation of standard generative models is that they are not explicitly optimized to maximize these scores; rather, they merely aim to approximate the distribution of known crystals.
Recent studies have begun using reinforcement learning (RL) to guide pretrained generative models in producing samples with target properties \citep{park2025guiding, xu2025plaid++, jia2024llmatdesign, cao2025crystalformer}.
In the RL framework, a reward function is first constructed to reflect the desired attributes, after which the model is updated to generate samples that yield high reward values.
Although a variety of reward functions have been employed in the literature, to date, no study has utilized the SUN score for this purpose. 
In this section, we investigate the potential of adopting SUN or cSUN as a reward signal.

\paragraph{Experimental Settings}
We adopt Chemeleon2 \citep{park2025guiding}, characterized by its latent diffusion architecture, as the backbone framework.
The training protocol for this model comprises three distinct stages: first, a Variational Autoencoder (VAE) \citep{kingma2013auto} is constructed to embed crystal structures into a continuous latent space; second, a diffusion model \citep{ho2020denoising} is trained within this latent space while keeping the VAE parameters fixed; and finally, the diffusion model is further updated using a RL algorithm called Group Relative Policy Optimization (GRPO) \citep{shao2024deepseekmath}.
By using a complex reward function consisting of validity, diversity, and creativity terms, the original authors successfully improved the $\overline{\mathrm{SUN}}_\mathrm{smat}$ score.
For our experiments, we exclusively modify the reward function, while maintaining the original dataset (MP20) and training configurations, such as the number of steps and learning rate.
Specifically, we employ either $\overline{\mathrm{SUN}}_\mathrm{smat}$ or $\overline{\mathrm{SUN}}_\mathrm{elm{+}am}$ as the reward signal.

\begin{figure}
    \centering
    \includegraphics[width=0.6\linewidth]{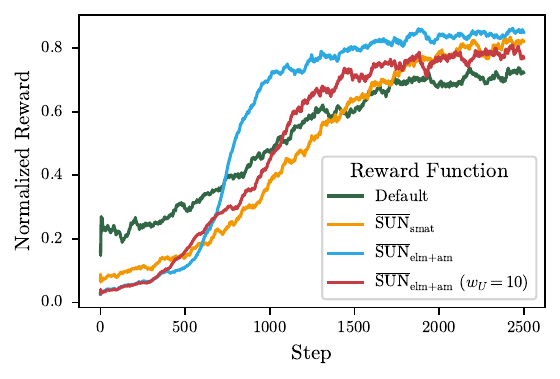}
    \caption{Training dynamics of Chemeleon2 under different reward function regimes for reinforcement learning. To make visualization easier, each reward value is normalized between 0 and 1 and smoothed out by taking the average of the last 50 steps at each time step. The model achieves effective convergence regardless of whether the reward signal is binary or continuous.}
    \label{fig:rl_train_dynamics}
\end{figure}

\paragraph{Impact of Reward Continuity on Optimization}
Initially, we hypothesized that $\overline{\mathrm{SUN}}_\mathrm{elm{+}am}$ would be easier to optimize than $\overline{\mathrm{SUN}}_\mathrm{smat}$ thanks to its continuous nature.
We anticipated not only faster convergence but also convergence to a superior local optimum, which would translate to a higher evaluation score even when assessed against the $\overline{\mathrm{SUN}}_\mathrm{smat}$ metric. (Recall the positive correlation observed between $\overline{\mathrm{SUN}}_\mathrm{smat}$ and $\overline{\mathrm{SUN}}_\mathrm{elm{+}am}$ in Figure~\ref{fig:sun_csun_mattergen}(b).)
However, the experimental results diverged from these expectations.
Figure~\ref{fig:rl_train_dynamics} illustrates how the reward value evolves during training.
Although RL with $\overline{\mathrm{SUN}}_\mathrm{elm{+}am}$ exhibits faster convergence compared to $\overline{\mathrm{SUN}}_\mathrm{smat}$, the disparity diminishes when the cSUN weights are adjusted (specifically, $w_U=10$). 
This trajectory suggests that Chemeleon2 possesses sufficient capacity to learn effectively from discrete reward signals, implying that continuous rewards do not guarantee accelerated convergence.
Table~\ref{tab:rl_score} compares the performance after RL with different reward functions.
Excluding the final column (to be discussed subsequently), RL with $\overline{\mathrm{SUN}}_\mathrm{smat}$ and $\overline{\mathrm{SUN}}_\mathrm{elm{+}am}$ yields maximal performance on $\overline{\mathrm{SUN}}_\mathrm{smat}$ and $\overline{\mathrm{SUN}}_\mathrm{elm{+}am}$, respectively. 
In essence, direct optimization of a specific metric results in the best performance on that same metric. 
Thus, it can be concluded that continuous reward functions do not necessarily steer the model toward a better local optimum.

\begin{table}
    \caption{Comparative assessment of $\overline{\mathrm{SUN}}_\mathrm{smat}$ and $\overline{\mathrm{SUN}}_\mathrm{elm{+}am}$ scores for Chemeleon2 under different reward signals for reinforcement learning. The highest score is in bold, and the second highest score is underlined. Notably, adjusting $w_U$ to 10 in $\overline{\mathrm{SUN}}_\mathrm{elm{+}am}$ significantly improves performance in both metrics compared to the default weights, indicating that high flexibility in the continuous reward can help optimization in this Chemeleon2/GRPO setting.}
    \label{tab:rl_score}
    \centering
    \begin{tabular}{@{}cccccc@{}}
        \toprule
        && \multicolumn{4}{c}{Reward function}\\
        \cmidrule(r){3-6}
       && default & $\overline{\mathrm{SUN}}_\mathrm{smat}$ & $\overline{\mathrm{SUN}}_\mathrm{elm{+}am}$ & $\overline{\mathrm{SUN}}_\mathrm{elm{+}am}$ ($w_U=10$) \\
        \midrule
        \multirow{2}{*}{Evaluation metric} & $\overline{\mathrm{SUN}}_\mathrm{smat}$ & 0.2651 & $\mathbf{0.4273}$ & 0.0478 & \underline{0.3946}\\
        & $\overline{\mathrm{SUN}}_\mathrm{elm{+}am}$ & 0.1701 & 0.3087 & \underline{0.5396} & $\mathbf{0.5643}$ \\
        \bottomrule
    \end{tabular}
\end{table}

\paragraph{Enhancing Uniqueness via Tunable cSUN}
To investigate in detail what kind of compounds are generated, we examined the number of unique compositions and the three most frequent compositions in each set of 10k structures sampled after RL (Table~\ref{tab:reward_hacking}). It is evident from the results that models after RL with either $\overline{\mathrm{SUN}}_\mathrm{smat}$ or $\overline{\mathrm{SUN}}_\mathrm{elm{+}am}$ focus on generating specific compositions. While it is natural that the sample distribution becomes biased toward a specific area of material space after RL, the generation of 900 or 611 novel, stable structures for the same composition is unrealistic.
A possible explanation for this phenomenon is the large discrepancy between the default batch size used during RL training (160) and the number of samples used for evaluation (10,000). 
During RL, the model is optimized to generate 160 samples that are as mutually unique as possible within each batch. 
However, when this 160-sample generation process is repeated many times, there is no guarantee that samples from different batches will remain dissimilar to one another. 
This also explains why the final reward value of $\overline{\mathrm{SUN}}_\mathrm{elm{+}am}$ in Figure~\ref{fig:rl_train_dynamics} is higher than the evaluation $\overline{\mathrm{SUN}}_\mathrm{elm{+}am}$ score in Table~\ref{tab:rl_score}. 
A straightforward solution would be to increase the batch size or to store previous batches and compare each new batch against them. 
Both approaches, however, can be computationally costly, as computing larger distance matrices at each step increases both memory usage and runtime. 
Therefore, we instead focus on enhancing sample diversity within each batch as much as possible. 
Notably, 900 occurrences of \ce{CsHg5} among 10,000 samples correspond to an average of 14.4 occurrences per batch, indicating substantial room for improvement. 
The proposed cSUN metric provides an intrinsic mechanism for promoting within-batch diversity by increasing the uniqueness weight, $w_U$.
Accordingly, we increased $w_U$ from 1 to 10 and analyzed the resulting structures.
The data presented in the final column of Table~\ref{tab:reward_hacking} indicate that this adjustment successfully alleviates compositional bias. 
Specifically, the model generates a 6.9-fold increase in unique compositions, while the count of samples belonging to the most frequent composition drops from 900 to 147. 
Although compositional uniqueness remains lower than the default setting in the original Chemeleon2 paper, adjusting the weights for $\mathrm{cSUN}_\mathrm{elm{+}am}$ demonstrably enhances compositional diversity.
Notably, setting $w_U = 10$ also facilitates convergence to a superior local optimum. 
As shown in Table~\ref{tab:rl_score}, RL with $\overline{\mathrm{SUN}}_\mathrm{elm{+}am}$ ($w_U = 10$) achieved a higher $\overline{\mathrm{SUN}}_\mathrm{elm{+}am}$ score (0.5643) than that obtained by direct optimization (0.5396). 
Furthermore, the $\overline{\mathrm{SUN}}_\mathrm{smat}$ score improved substantially from the default weight baseline (0.0478 $\to$ 0.3946), approaching the peak score (0.4273) achieved via direct optimization.
To further improve sample uniqueness, it is important to promote inter-batch uniqueness in addition to intra-batch uniqueness.
Finally, we emphasize that we do not claim that cSUN generally guarantees convergence to better optima, as this experiment was conducted using a single backbone model, a single RL algorithm, and a single dataset.

\begin{table}
    \caption{Number of unique compositions and the top 3 most frequent compositions in the 10k samples from Chemeleon2. Each column corresponds to a different reward function for reinforcement learning. Reinforcement learning with $\overline{\mathrm{SUN}}_\mathrm{smat}$ or $\overline{\mathrm{SUN}}_\mathrm{elm{+}am}$ results in extremely focusing on specific compositions. Enhancing diversity by setting $w_U$ to 10 in $\overline{\mathrm{SUN}}_\mathrm{elm{+}am}$ mitigates the issue.}
    \label{tab:reward_hacking}
    \centering
    \begin{tabular}{@{}ccccc@{}}
        \toprule
        & \multicolumn{4}{c}{Reward function}\\
        \cmidrule(r){2-5}
       & default & $\overline{\mathrm{SUN}}_\mathrm{smat}$ & $\overline{\mathrm{SUN}}_\mathrm{elm{+}am}$ & $\overline{\mathrm{SUN}}_\mathrm{elm{+}am}$ ($w_U=10$) \\
        \midrule
        Number of unique compositions & 6807 & 4067 & 287 & 1980\\
        Most frequent composition & \ce{K} (37) & \ce{Gd5Ir2} (611) & \ce{CsHg5} (900) & \ce{AcScPd2} (147)\\
        Second frequent composition & \ce{Li3Co4O8} (22) & \ce{Ac4Pt} (141) & \ce{AcScHg2} (750) & \ce{AcPb2Bi} (145) \\
        Third frequent composition & \ce{TbZnGa} (21) & \ce{Ac12Ni6Pb} (86) & \ce{Cs6RbHg} (743) & \ce{AcScPbBi} (124) \\
        \bottomrule
    \end{tabular}
\end{table}

\section{Conclusion}
We have addressed limitations of conventional binary uniqueness, novelty, and stability metrics for generative materials models by introducing continuous counterparts for each.
We further integrate them into a single metric termed cSUN, which overcomes the coarseness of the binary classification inherent in the original SUN metric.
In addition to the capability of offering detailed rankings of candidates and the flexibility of prioritizing some components over others, our metrics demonstrate theoretical advantages, such as robustness against small atomic shifts and invariance to sample permutation.
Experimental results confirm that our metrics provide granular insights into the sample distribution and help identify the most promising candidates.
Moreover, cSUN is shown to be utilized as an effective reward signal in reinforcement learning, where its adjustable weights effectively enhance sample diversity and convergence to a better local optima in our tested setting.
A promising avenue for future research involves employing learnable distance functions for cU and cN evaluations.
Specifically, investigating distances derived from machine learning foundation models (e.g., universal force fields) warrants attention, as these models are expected to capture rich chemical similarities learned from vast material databases.
Additionally, future work could explore more sophisticated formulations for cS.
Although we examined three different functional forms of cS in \ref*{appendix:sec:sec:cs_sensitivity}, all of them are composition-independent.
A natural extension would be to make cS composition-dependent, since some compositions are more likely to be synthesizable than others under the same $E_\mathrm{hull}$ value \citep{sun2016thermodynamic}.
Finally, the accuracy of MACE-MPA-0 energy predictions could be further improved. 
Although the model has been reported to show good agreement with DFT formation energies, it has also been reported to systematically underpredict formation energies \citep{batatia2025foundation}. 
Therefore, systematic calibration of the predicted energies before computing cS may enable more rigorous stability evaluation.
We hope that our metrics will serve as a standard for evaluation and as effective reward signals to guide the continued advancement of generative models in this domain.

\ack{We would like to thank Vitaliy Kurlin (University of Liverpool) for valuable feedback on the theoretical analysis of crystal distance functions.}

\funding{We acknowledge support of the AIchemy Hub (EPSRC grant EP/Y028775/1 and EP/Y028759/1), as well as an Imperial College President's PhD scholarship .}

\conflict{A. W. is Chief Scientific Officer at CuspAI.}

\roles{Masahiro Negishi: conceptualization, data curation, formal analysis, investigation, methodology, software, validation, visualization, writing (original draft), and writing (review and editing). Hyunsoo Park: supervision and writing (review and editing). Kinga O. Mastej: visualization and writing (review and editing). Aron Walsh: conceptualization, funding acquisition, project administration, resources, supervision, writing (review and editing).}

\data{The open-source Python package implementing the proposed metrics and the code required to reproduce the results is available at \url{https://github.com/WMD-group/xtalmet}. The crystal structures generated by the models evaluated in this study are hosted at \url{https://huggingface.co/datasets/masahiro-negishi/xtalmet}. This study utilized exclusively publicly available datasets.}

\suppdata{Supporting Information. (336KB PDF): Detailed definitions and theoretical analyses of crystal distance functions, as well as detailed experimental settings and additional results.}

\bibliographystyle{iopart-num-long}
\bibliography{reference}

\newpage

\begin{center}
    \vspace*{2cm}
    {\Huge \textbf{Supporting Information}} \\
    \vspace{1cm}
\end{center}

\etocdepthtag.toc{SI}
\etocsettagdepth{mt}{none}
\etocsettagdepth{SI}{subsection}
\etocsettocstyle{}{} 

\setcounter{section}{0}
\setcounter{page}{1}
\setcounter{figure}{0}
\setcounter{table}{0}
\setcounter{equation}{0}
\setcounter{theorem}{0}

\renewcommand{\thesection}{S\arabic{section}}
\renewcommand{\thefigure}{S\arabic{figure}}
\renewcommand{\thetable}{S\arabic{table}}
\renewcommand{\theequation}{S\arabic{equation}}
\renewcommand{\thetheorem}{S\arabic{theorem}}

\renewcommand{\theHsection}{SI.\thesection}
\renewcommand{\theHfigure}{SI.\thefigure}
\renewcommand{\theHtable}{SI.\thetable}
\renewcommand{\theHequation}{SI.\theequation}
\renewcommand{\theHtheorem}{SI.\thetheorem}

\makeatletter
\renewcommand{\citenumfont}[1]{S#1}
\renewcommand{\bibnumfmt}[1]{[S#1]}
\makeatother

This file contains detailed definitions and theoretical analyses of crystal distance functions, as well as detailed experimental settings and additional results.

\tableofcontents

\section{Crystal Distance Functions}

\ref{appendix:sec:sec:distance_definition} establishes the formal definitions of the crystal distance functions utilized throughout this study.
Subsequently, Section~\ref{appendix:sec:sec:other_distances} outlines the rationale for selecting $d_\mathrm{elm}$, $d_\mathrm{am}$, and $d_\mathrm{elm{+}am}$ as the preferred continuous metrics, distinguishing them from alternative approaches.
Finally, \ref{appendix:sec:sec:invariance} provides a mathematical proof demonstrating that the model-level uniqueness score, $\overline{\mathrm{U}}$, remains invariant to sample permutation, provided that the underlying distance $d$ is continuous or constitutes a true pseudometric. 
This result implies that all distance metrics introduced in this work satisfy the invariance requirement, with the notable exception of $d_\mathrm{smat}$.

\subsection{Definitions}
\label{appendix:sec:sec:distance_definition}

\paragraph{\bm{$d_\mathrm{smat}$}} 
is the most widely used discrete distance, derived from the ``fit" method of the StructureMatcher class in the pymatgen library \citepsi{si-ong2013python}. 
The algorithm operates via a two-step procedure:
\begin{enumerate}
    \item Compositional difference: The function first checks for compositional identity. Distinct compositions yield a distance of 1.
    \item Structural difference: Given a compositional match, the algorithm reduces the input structures to their primitive cells. Subsequently, it systematically searches through all possible lattice transformations and translations to align one structure with the other.  A match is declared (distance = 0) if an alignment is found where all corresponding atoms are of the same chemical element and the distance between each matched atom pair falls below a predefined threshold. Otherwise, the function returns a distance of 1.
\end{enumerate}
When applied to the examples in Table~\ref*{tab:example} (main text), this function returns a distance of 1 for every pair except for the first one (wz-ZnO and its supercell), which is compositionally and structurally identical.
It is important to note that $d_\mathrm{smat}$ is mathematically not a pseudometric because it can violate the triangle inequality. 
To illustrate, consider three crystals, $x, x'$, and $x''$. 
Deviations between $x$ and $x'$, and between $x'$ and $x''$, may fall within the matching threshold (implying $d_\mathrm{smat}(x, x') = 0$ and $d_\mathrm{smat}(x', x'') = 0$).
However, the cumulative deviation may result in $d_\mathrm{smat}(x, x'') = 1$, thereby violating the condition ($0 + 0 \ngeq 1$). 
Nevertheless, for simplicity, we refer to $d_\mathrm{smat}$ as a ``distance” throughout this work, as we do with other pseudometric functions introduced below.

\paragraph{\bm{$d_\mathrm{comp}$}}
checks whether the compositions of two input structures are identical: 
\begin{equation*}
    d_\mathrm{comp}(x_1, x_2) \coloneqq
    \begin{cases*}
        0 & if the compositions are the same,\\
        1 & otherwise.
    \end{cases*}
\end{equation*}
Applying this definition to the examples in Table~\ref*{tab:example} (main text), $d_\mathrm{comp}$ yields a value of 0 for the first two pairs, as each pair shares the same composition (ZnO).

\paragraph{\bm{$d_\mathrm{wyckoff}$}}
is a discrete metric that quantifies structural dissimilarity based on space group symmetry and Wyckoff positions.
We offer a brief overview of these concepts below. 
More comprehensive information can be found in standard crystallography literature, such as \citepsi{si-sands1993introduction}.

A space group provides a full description of a crystal's symmetry. 
It encompasses the set of all possible symmetry operations (such as translations, rotations, and reflections) that leave the crystal unchanged. 
Crystals are classified into one of 230 unique space groups.
For instance, wz-ZnO in Figure~\ref*{fig:example}(a) (main text) has symmetry operations such as a six-fold screw axis along the c-axis, and is assigned to space group 186.

Within a given space group, atomic positions are categorized by Wyckoff positions. 
A Wyckoff position serves as a label for a set of symmetrically equivalent points generated by the space group's operations. 
Consequently, placing an atom at a specific Wyckoff site implicitly determines the coordinates of all equivalent atoms. 
Taking wz-ZnO as an example, the unit cell contains two Zn and two O atoms. 
Both species independently occupy the Wyckoff position ``$b$'' (representative coordinate: $(\frac{1}{3}, \frac{2}{3}, z)$). 
Applying the symmetry operations of space group 186 generates the symmetry-equivalent position at $(\frac{2}{3}, \frac{1}{3}, z + \frac{1}{2})$. 
Thus, the structure is fully defined by specifying the Wyckoff position and the free parameter $z$ (approximately 0 for Zn and 0.38 for O) for a single representative atom.

In summary, a crystal can be fully described by its space group, lattice parameters (the lengths and angles of the three lattice vectors), the occupied Wyckoff positions, and the chemical elements at those positions. 
For instance, wz-ZnO in Figure~\ref*{fig:example}(a) (main text) can be represented as follows:
\begin{align*}
    &\text{space group} : 186\\
    &\text{lattice parameters} : (3.24 \text{\AA}, \ 3.24 \text{\AA}, \ 5.22 \text{\AA}, \ 90^\circ, \ 90^\circ, \ 120^\circ)\\
    &\text{Wyckoff positions and elements} : \text{Zn @ }b \ (z=0), \text{O @ } b \ (z=0.38).
\end{align*}
$d_\mathrm{wyckoff}$ evaluates whether two crystals share the same space group and the same multiset of occupied Wyckoff letters:
\begin{equation*}
    d_\mathrm{wyckoff}(x_1, x_2) \coloneqq
    \begin{cases*}
        0 & if they have the same space group and the Wyckoff letters,\\
        1 & otherwise.
    \end{cases*}
\end{equation*}
In the case of wz-ZnO, the metric only uses the information of the space group (186), and the multiset of Wyckoff letters, $\{\!\{b, b\}\!\}$.
It abstracts away compositional details and specific atomic coordinates.
Regarding the examples in Table~\ref*{tab:example} (main text), both the first and third pairs have the same space group and Wyckoff letters, so their $d_\mathrm{wyckoff}$ distance is 0. 
Note that the exact coordinates of atoms are different for the third pair.

\paragraph{\bm{$d_\mathrm{elm}$}}
allows for continuous measurement of compositional distance. 
It treats composition as a histogram of elements and measures the optimal transport cost of moving one histogram to another \citepsi{si-hargreaves2020earth}.
Formally, consider two structures $x_1$ and $x_2$ containing $K_1$ and $K_2$ distinct elements, respectively.
Let $n_{1, i}$ denote the number of atoms for element $i$ in $x_1$, and $n_{2, j}$ denote the count for element $j$ in $x_2$. 
The total number of atoms is given by $n_1 \coloneqq \sum_{i=1}^{K_1} n_{1, i}$ and $n_2 \coloneqq \sum_{j=1}^{K_2} n_{2, j}$.
Then, the original unnormalized Element Mover's Distance between these two crystals is defined as:
\begin{equation*}
    \tilde{d}_\mathrm{elm}(x_1, x_2) \coloneqq \min_{P \in \mathbb{R}_{\ge0}^{K_1 \times K_2}} \sum_{i=1}^{K_1} \sum_{j=1}^{K_2} C_{ij} P_{ij} \ \text{ s.t. } \sum_{j=1}^{K_2} P_{ij} = \frac{n_{1,i}}{n_1} \text{ and } \sum_{i=1}^{K_1} P_{ij} = \frac{n_{2,j}}{n_2}.
\end{equation*}
Here, $P_{ij}$ represents the transport plan (or flow) from element $i$ in $x_1$ to element $j$ in $x_2$.
The constraints ensure the conservation of mass, guaranteeing that the normalized abundance of each element in $x_1$ is fully mapped to the distribution of $x_2$.
The term $C_{ij}$ denotes the ground cost matrix, representing the chemical dissimilarity between elements $i$ and $j$. 
Intuitively, $C_{ij}$ takes a small value when elements possess similar properties.
This study used $C_{ij}$ derived from a one-dimensional periodic table, which reflects the statistical likelihood that element $i$ can be replaced by $j$ in an experimental database \citepsi{si-glawe2016optimal}.
Notably, the one-dimensional nature of this cost function allows faster computation of the seemingly complex $\tilde{d}_\mathrm{elm}$ in linear time relative to the number of distinct elements.
The normalized $d_\mathrm{elm}$ is defined via $d_\mathrm{elm} \coloneqq \frac{\tilde{d}_\mathrm{elm}}{1+\tilde{d}_\mathrm{elm}}$.
Consistent with $d_\mathrm{comp}$, $d_\mathrm{elm}$ returns a distance of 0 for the first two pairs in Table~\ref*{tab:example} (main text), because $d_\mathrm{elm}$ depends solely on composition. 
However, the advantage of a continuous metric becomes clear in the third and fourth pairs. 
For these examples, $d_\mathrm{elm}$ provides a more nuanced comparison of compositional similarity, whereas the binary $d_\mathrm{comp}$ can only indicate that they are different.

\paragraph{\bm{$d_\mathrm{am}$}}
is a continuous structural distance defined as the $L_\infty$ distance between structural fingerprints, called Average Minimum Distance (AMD) vectors \citepsi{si-widdowson2022average}.
The original unnormalized distance is defined as:
\begin{equation*}
    \tilde{d}_\mathrm{am}(x_1, x_2) \coloneqq \|\mathrm{AMD}(x_1) - \mathrm{AMD}(x_2)\|_\infty,
\end{equation*}
where $\mathrm{AMD}(x)$ denotes the AMD vector of $x$.
The $k^\mathrm{th}$ element of an AMD vector stands for the average distance from an atom to its $k^\mathrm{th}$ nearest neighbor, with the average taken over all atoms in the structure.
The normalized $d_\mathrm{am}$ is defined via $d_\mathrm{am} \coloneqq \frac{\tilde{d}_\mathrm{am}}{1+\tilde{d}_\mathrm{am}}$.
Applying this metric to Table~\ref*{tab:example} (main text), $d_\mathrm{am}$ yields a value of zero for the first pair, reflecting the structural identity between the primitive cell and its supercell.
For the remaining three pairs, $d_\mathrm{am}$ returns non-zero values. 
Of these three pairs, (wz-ZnO, wz-GaN) exhibits the minimum distance. 
This result is consistent with the fact that only this pair shares the same space group and Wyckoff letters.
Compared with $d_\mathrm{wyckoff}$, $d_\mathrm{am}$ does not discard the information about exact coordinates and serves as a more detailed indicator of structural similarity.

\subsection{Other Continuous Distances}
\label{appendix:sec:sec:other_distances}
For continuous, compositional comparison, several studies \citepsi{si-nguyen2023hierarchical, si-gruver2023llmtime, si-mohanty2025crystext, si-gan2025matllmsearchcrystalstructurediscovery} have employed the Euclidean distance between Magpie fingerprints, which consist of 145 attributes, including stoichiometric attributes and elemental property statistics \citepsi{si-ward2016general}.
However, we opted not to use this metric because certain dimensions of the fingerprint lack distinct semantic significance.
For instance, one dimension corresponds to the average atomic number of the constituent elements.
Nevertheless, it is worth noting that, in principle, any continuous compositional distance satisfying the three requirements listed in Table~\ref*{tab:property} (main text) is applicable to uniqueness and novelty evaluation.
As a continuous structural distance, the Euclidean distance between structural CrystalNN fingerprints \citepsi{si-zimmermann2020local} has been widely utilized \citepsi{si-nguyen2023hierarchical, si-gruver2023llmtime, si-mohanty2025crystext, si-gan2025matllmsearchcrystalstructurediscovery, si-ren2022invertible, si-govindarajan2024learning, si-vasylenko2025physics}.
We decided not to use it because it fails to satisfy the Lipschitz continuity.
A similar limitation applies to another commonly used continuous structural distance: the regularized-entropy match distance using structural Smooth Overlap of Atomic Positions (SOAPs) descriptors \citepsi{si-de2016comparing}.
Finally, to ensure high interpretability, we refrain from employing distances between learned features, such as embeddings from foundation machine learning force fields.
Extending this framework to include such embeddings would be an interesting direction for future work.

\subsection{Invariance of Uniqueness Against Permutation of Generated Samples}
\label{appendix:sec:sec:invariance}
As discussed in Section~\ref*{sec:sec:distance_theory} (main text), $\overline{\mathrm{U}}_\mathrm{smat}$ lacks invariance to the generation order due to the violation of the triangle inequality by $d_\mathrm{smat}$. 
Here, we provide a formal proof demonstrating that $\overline{\mathrm{U}}$ derived from a true discrete pseudometric is inherently invariant.
Note that $\overline{\mathrm{U}}$ using a continuous distance is always independent of the generation order because it is simply the average distance between generated samples.
\begin{theorem}[Permutation Invariance of $\overline{\mathrm{U}}$ with Discrete Pseudometric]
    Let $d_\mathrm{discrete}$ be a discrete pseudometric taking values in $\{0, 1\}$. 
    Suppose it satisfies the axioms of a pseudometric for any $x, x', x''$:
    \begin{align*}
        &d_\mathrm{discrete}(x, x) = 0,\\
        &d_\mathrm{discrete}(x, x') = d_\mathrm{discrete}(x', x),\\
        &d_\mathrm{discrete}(x, x'') \le d_\mathrm{discrete}(x, x') + d_\mathrm{discrete}(x', x'').
    \end{align*}
    Then, the model-level uniqueness $\overline{\mathrm{U}}_\mathrm{discrete}$, which is an average of sample-level $\mathrm{U}_\mathrm{discrete}$ in Equation~\ref*{eq:sample-level-u} (main text), is invariant under any permutation of the generated samples.
\end{theorem}
\begin{proof}
    We define a relation $\sim$ on the set of generated samples such that $x \sim x'$ if and only if $d_\mathrm{discrete}(x, x') = 0$.
    Using the properties of the pseudometric, we show that $\sim$ is an equivalence relation:
    \begin{enumerate}
        \item Reflexivity: $d_\mathrm{discrete}(x, x) = 0 \implies x \sim x$.
        \item Symmetry: $x \sim x' \implies d_\mathrm{discrete}(x, x') = 0 \implies d_\mathrm{discrete}(x', x) = 0 \implies x' \sim x$.
        \item Transitivity: If $x \sim x'$ and $x' \sim x''$, then $d_\mathrm{discrete}(x, x') = 0$ and $d_\mathrm{discrete}(x', x'') = 0$. By the triangle inequality:
        \begin{equation*}
            d_\mathrm{discrete}(x, x'') \le d_\mathrm{discrete}(x, x') + d_\mathrm{discrete}(x', x'') = 0 + 0 = 0.
        \end{equation*}
        Since distance is non-negative, $d_\mathrm{discrete}(x, x'') = 0$, implying $x \sim x''$.
    \end{enumerate}
    Since $\sim$ is an equivalence relation, it partitions the samples into a unique set of disjoint equivalence classes, denoted as $\{S_1, S_2, \ldots, S_m\}$.
    The calculation of $\overline{\mathrm{U}}_\mathrm{discrete}$ is equivalent to counting the number of these equivalence classes.
    Because the set of equivalence classes is intrinsic to the dataset and independent of the order in which elements are listed, the value $\frac{m}{n}$ is invariant to permutation.
\end{proof}

\section{Experiments}

\subsection{Experimental Settings}
In this study, we evaluated 10,000 structures generated by each of seven distinct generative models: CDVAE \citepsi{si-xie2022crystal}, DiffCSP \citepsi{si-jiao2023crystal}, DiffCSP++ \citepsi{si-jiao2024space}, MatterGen \citepsi{si-MatterGen2025}, Chemeleon-DNG \citepsi{si-park2025exploration}, ADiT \citepsi{si-joshi2025allatom}, and Chemeleon2 \citepsi{si-park2025guiding}
All models were trained on the MP20 dataset and subsequently employed for de novo crystal sampling.
Regarding data acquisition, samples for CDVAE, DiffCSP, DiffCSP++, and MatterGen were generated using the official codebases and pre-trained checkpoints.
For Chemeleon-DNG, ADiT, and Chemeleon2, we utilized the publicly available sets of 10k structures released by the authors.
Novelty was benchmarked against the MP20 training set. 
To quantify thermodynamic stability, the formation energy of each candidate was first predicted via the MACE-MPA-0 machine learning force field \citepsi{si-batatia2025foundation}, followed by the computation of the energy above the convex hull ($E_\mathrm{hull}$) relative to the Materials Project database \citepsi{si-jain2013commentary}.

\subsection{Actual Values of $\overline{\mathrm{U}}$, $\overline{\mathrm{N}}$, and $\overline{\mathrm{SUN}}$ scores}

\begin{table}
    \caption{The uniqueness scores $\overline{\mathrm{U}}$ for each generative model based on different distances. The highest score in each row is in bold, and the second highest score is underlined. Note that the MP20 Test set is included as a baseline and is excluded from the ranking.}
    \label{appendix:tab:uni}
    \centering
    \begin{tabular}{lllllllll}
        \toprule
        & CDVAE & DiffCSP & DiffCSP++ & MatterGen & Chemeleon-DNG & ADiT & Chemeleon2 & MP20 Test\\
        \midrule
        $d_\mathrm{comp}$ & $\mathbf{0.9717}$ & 0.9461 & 0.9522 & \underline{0.9523} & 0.9373 & 0.7743 & 0.6807 & 0.9434\\
        $d_\mathrm{elm}$ & 0.9379 & $\mathbf{0.9446}$ & 0.9440 & \underline{0.9441} & 0.9440 & 0.9402 & 0.9341 & 0.9439\\
        $d_\mathrm{wyckoff}$ & 0.0045 & 0.0326 & $\mathbf{0.1910}$ & 0.0470 & \underline{0.0640} & 0.0038 & 0.0039 & 0.2125\\
        $d_\mathrm{am}$ & 0.5032 & 0.5435 & 0.5389 & \underline{0.5445} & $\mathbf{0.5528}$ & 0.5131 & 0.5384 & 0.5481\\
        $d_\mathrm{smat}$ & $\mathbf{0.9945}$ & 0.9771 & 0.9806 & \underline{0.9838} & 0.9791 & 0.8841 & 0.7936 & 0.9940\\
        $d_\mathrm{elm{+}am}$ & 0.8406 & \underline{0.8549} & 0.8534 & 0.8547 & $\mathbf{0.8565}$ & 0.8447 & 0.8456 & 0.8553\\
        \bottomrule
    \end{tabular}
\end{table}

\begin{table}
    \caption{The novelty scores $\overline{\mathrm{N}}$ for each generative model based on different distances. The highest score in each row is in bold, and the second highest score is underlined. Note that the MP20 Test set is included as a baseline and is excluded from the ranking.}
    \label{appendix:tab:nov}
    \centering
    \begin{tabular}{lllllllll}
        \toprule
        & CDVAE & DiffCSP & DiffCSP++ & MatterGen & Chemeleon-DNG & ADiT & Chemeleon2 & MP20 Test\\
        \midrule
        $d_\mathrm{comp}$ & $\mathbf{0.9221}$ & \underline{0.8207} & 0.8121 & 0.8156 & 0.7497 & 0.2352 & 0.8090 & 0.7904\\
        $d_\mathrm{elm}$ & $\mathbf{0.5230}$ & \underline{0.3954} & 0.3951 & 0.3883 & 0.3306 & 0.0655 & 0.3348 & 0.1973\\
        $d_\mathrm{wyckoff}$ & 0.0352 & 0.0814 & 0.0098 & \underline{0.0967} & $\mathbf{0.1137}$ & 0.0425 & 0.0371 & 0.0500\\
        $d_\mathrm{am}$ & $\mathbf{0.1627}$ & 0.1095 & 0.1082 & 0.1199 & 0.1073 & 0.1176 & \underline{0.1367} & 0.0705\\
        $d_\mathrm{smat}$ & $\mathbf{0.9892}$ & 0.9050 & 0.8925 & 0.9158 & 0.8630 & 0.4582 & \underline{0.9731} & 0.9608\\
        $d_\mathrm{elm{+}am}$ & $\mathbf{0.4920}$ & 0.3717 & \underline{0.3725} & 0.3681 & 0.3150 & 0.0923 & 0.3302 & 0.1913\\
        \bottomrule
    \end{tabular}
\end{table}

\begin{table}
    \caption{The $\overline{\mathrm{SUN}}$ scores for each generative model based on different distances. The highest score in each row is in bold, and the second highest score is underlined. Note that the MP20 Test set is included as a baseline and is excluded from the ranking.}
    \label{appendix:tab:sun}
    \centering
    \begin{tabular}{lllllllll}
        \toprule
        & CDVAE & DiffCSP & DiffCSP++ & MatterGen & Chemeleon-DNG & ADiT & Chemeleon2 & MP20 Test\\
        \midrule
        $d_\mathrm{comp}$ & 0.0290 & 0.1741 & 0.1551 & $\mathbf{0.2234}$ & \underline{0.2078} & 0.0565 & 0.1932 & 0.7300\\
        $d_\mathrm{elm}$ & 0.1188 & 0.1371 & 0.0938 & \underline{0.1891} & 0.1410 & 0.0178 & $\mathbf{0.1896}$ & 0.1714\\
        $d_\mathrm{wyckoff}$ & 0.0000 & 0.0071 & 0.0010 & $\mathbf{0.0149}$ & \underline{0.0138} & 0.0012 & 0.0011 & 0.0420\\
        $d_\mathrm{am}$ & 0.0209 & 0.0227 & 0.0152 & \underline{0.0362} & 0.0255 & 0.0272 & $\mathbf{0.0476}$ & 0.0358\\
        $d_\mathrm{smat}$ & 0.0318 & 0.2106 & 0.1820 & $\mathbf{0.2800}$ & 0.2562 & 0.0680 & \underline{0.2651} & 0.9021\\
        $d_\mathrm{elm{+}am}$ & 0.1003 & 0.1191 & 0.0828 & \underline{0.1640} & 0.1223 & 0.0259 & $\mathbf{0.1701}$ & 0.1503\\
        \bottomrule
    \end{tabular}
\end{table}

Figures~\ref*{fig:uni}, \ref*{fig:nov}, and \ref*{fig:sun} in the main text show the $\overline{\mathrm{U}}$, $\overline{\mathrm{N}}$, and $\overline{\mathrm{SUN}}$ scores of each model in the form of strip plots for easier understanding. 
Here, we present actual numerical scores in Tables~\ref{appendix:tab:uni}, \ref{appendix:tab:nov}, and \ref{appendix:tab:sun}.

\subsection{Sensitivity Analysis for Length of AMD Vector}
\label{appendix:sec:sec:am_k}

\begin{figure}
    \centering
    \includegraphics[width=\linewidth]{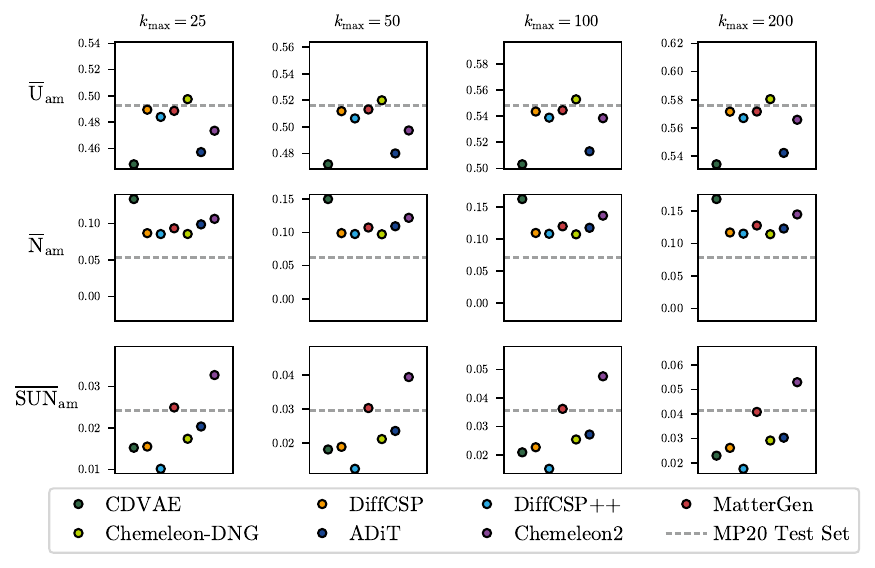}
    \caption{$\overline{\mathrm{U}}_\mathrm{am}$, $\overline{\mathrm{N}}_\mathrm{am}$, and $\overline{\mathrm{SUN}}_\mathrm{am}$ scores for each generative model under different AMD vector lengths $k\mathrm{max}$. Evaluation is performed using 10k unconditioned samples. The horizontal gray dashed line indicates the corresponding score for the MP20 test set. Although the absolute scores gradually increase with $k_\mathrm{max}$, the distributional patterns remain nearly identical within each row.}
    \label{appendix:fig:am_k_stripplot}
\end{figure}

\begin{figure}
    \centering
    \includegraphics[width=\linewidth]{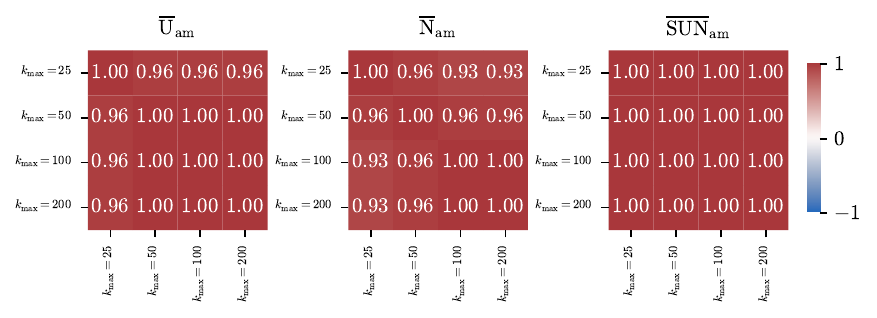}
    \caption{Spearman rank correlations between model scores obtained using different AMD vector lengths $k_\mathrm{max}$. All pairs show strong positive correlations, indicating that the model rankings for each score are robust to the choice of $k_\mathrm{max}$.}
    \label{appendix:fig:am_k_heatmap}
\end{figure}

This study defines a continuous structural metric using the $L_\infty$ distance between Average Minimum Distance (AMD) vectors. In this definition, the length of the AMD fingerprint, $k_\mathrm{max}$, is a hyperparameter (see~\ref{appendix:sec:sec:distance_definition}). The parameter $k_\mathrm{max}$ determines the number of neighboring atoms considered when constructing the structural fingerprint. Following the original work, we use the default value $k_\mathrm{max}=100$, meaning that distances up to the $100^\mathrm{th}$ nearest neighbor are considered for each atom. Here, we investigate how varying $k_\mathrm{max}$ affects the evaluations of $\overline{\mathrm{U}}_\mathrm{am}$, $\overline{\mathrm{N}}_\mathrm{am}$, and $\overline{\mathrm{SUN}}_\mathrm{am}$.

Figure~\ref{appendix:fig:am_k_stripplot} shows the $\overline{\mathrm{U}}_\mathrm{am}$, $\overline{\mathrm{N}}_\mathrm{am}$, and $\overline{\mathrm{SUN}}_\mathrm{am}$ scores for each model, computed using different AMD vector lengths $k_\mathrm{max}$. Because the $L_\infty$ distance between two vectors monotonically increases as additional dimensions are included, the absolute score values gradually increase with $k_\mathrm{max}$. However, the shapes of the score distributions are nearly identical across different values of $k_\mathrm{max}$. This observation is quantitatively supported by Figure~\ref{appendix:fig:am_k_heatmap}, which shows strong rank correlations between all pairs of $k_\mathrm{max}$ values. These results indicate that the proposed metrics are highly robust to the choice of $k_\mathrm{max}$.

\subsection{Sensitivity Analysis for Distance Normalization}
\label{appendix:sec:sec:normalize}

\begin{figure}
    \centering
    \includegraphics[width=\linewidth]{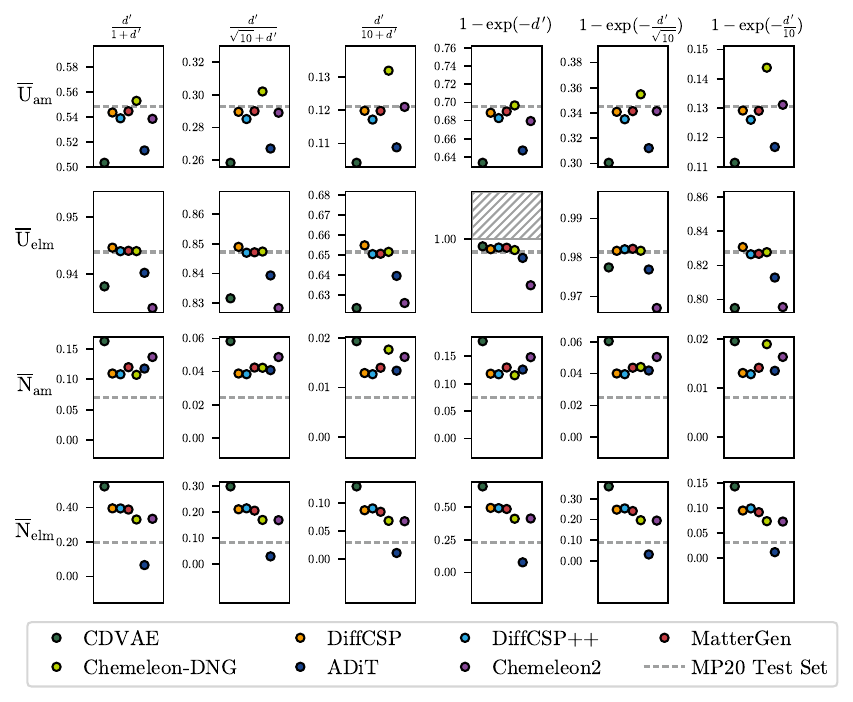}
    \caption{$\overline{\mathrm{U}}_\mathrm{am}$, $\overline{\mathrm{U}}_\mathrm{elm}$, $\overline{\mathrm{N}}_\mathrm{am}$, and $\overline{\mathrm{N}}_\mathrm{elm}$ scores for each generative model under different normalization methods for the distance functions. Evaluation is performed using 10k unconditioned samples. The horizontal gray dashed line indicates the corresponding score for the MP20 test set. Because $\overline{\mathrm{U}}$ and $\overline{\mathrm{N}}$ are bounded above by 1.0, regions where $\overline{\mathrm{U}}>1$ or $\overline{\mathrm{N}}>1$ are shaded in each subplot. Although the absolute scores vary with the normalization method, the model ranking trends are largely preserved.}
    \label{appendix:fig:normalize_stripplot}
\end{figure}

In the original formulations, the ranges of $d_\mathrm{elm}$ and $d_\mathrm{am}$ are $[0,\infty)$. In this work, we normalize these distances to $[0,1]$ so that the uniqueness and novelty metrics derived from them have the same range as those based on conventional binary distance functions. To this end, we use the transformation $d = \frac{d'}{1+d'}$, where $d'$ denotes the original unnormalized distance. This normalized distance is mathematically guaranteed to preserve the properties of a proper pseudometric. However, alternative normalization schemes may also preserve these pseudometric properties. In this section, we investigate how the choice of normalization method affects the evaluation of uniqueness and novelty.

\begin{figure}
    \centering
    \includegraphics[width=\linewidth]{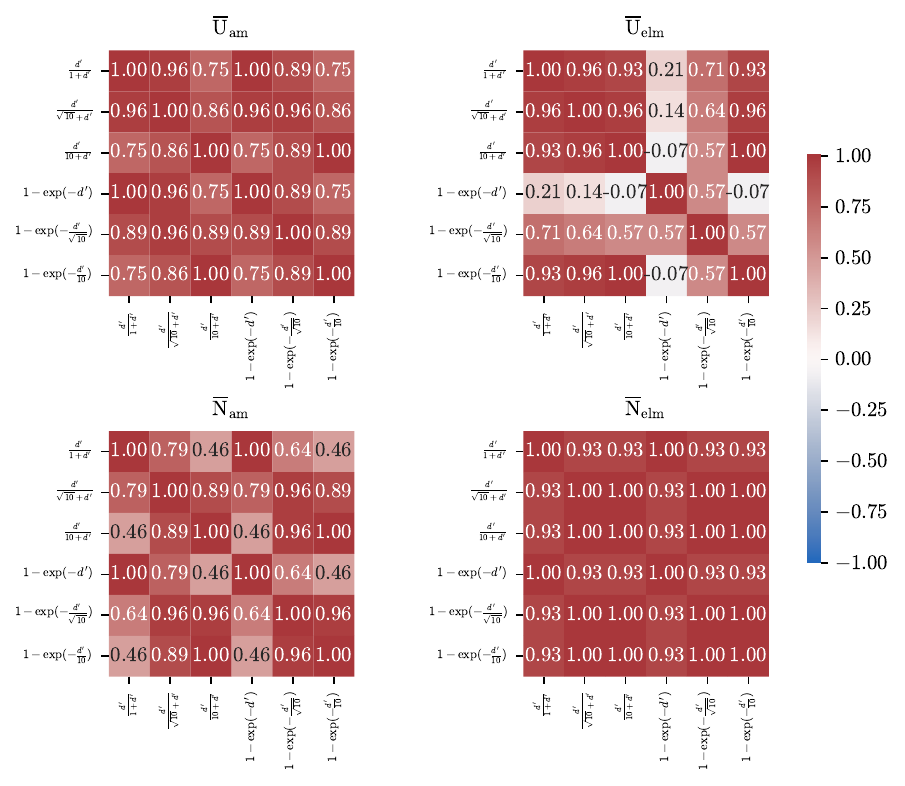}
    \caption{Spearman rank correlations between model scores obtained under different distance-normalization methods. Most pairs show positive correlations, indicating that the model rankings for each score are robust to the choice of normalization method, except for $d = 1 - \exp(-d')$ in $\overline{\mathrm{U}}_\mathrm{elm}$.}
    \label{appendix:fig:normalize_heatmap}
\end{figure}

Specifically, we compare six normalization schemes. The first three are fractional normalizations, defined as $d = \frac{d'}{c+d'}$ with $c = 1, \sqrt{10}, 10$, where $c=1$ corresponds to our default normalization. The remaining three are exponential normalizations, defined as $d = 1 - \exp(-d'/\lambda)$ with $\lambda = 1, \sqrt{10}, 10$. Figure~\ref{appendix:fig:normalize_stripplot} shows the uniqueness $\overline{\mathrm{U}}$ and novelty $\overline{\mathrm{N}}$ scores of different models under these six normalization methods. As expected, the absolute score values depend on the normalization method; however, the main ranking trends are largely preserved. This consistency is observed not only across different hyperparameter values of $c$ and $\lambda$, but also between the fractional and exponential normalizations. This observation is quantitatively supported by Figure~\ref{appendix:fig:normalize_heatmap}, which shows the Spearman rank correlations between pairs of normalization methods. Most pairs exhibit positive correlations, with the notable exception of $d = 1 - \exp(-d')$ for $\overline{\mathrm{U}}_\mathrm{elm}$. This exception arises because CDVAE achieves the highest $\overline{\mathrm{U}}_\mathrm{elm}$ under this normalization, whereas it ranks fifth or lower under the other normalizations. As discussed in Section~\ref*{sec:sec:un_experiment} in the main text, although CDVAE produces few exact compositional duplicates, its overall distribution in the continuous $d_\mathrm{elm}$ space is highly concentrated rather than dispersed. Because $d = 1 - \exp(-d')$ increases more rapidly with $d'$ than the other normalization methods, this concentrated distribution is not sufficiently penalized under this transformation. Indeed, in the limits $c \to 0$ and $\lambda \to 0$, the corresponding normalized distances become equivalent to the binary distance $d_\mathrm{comp}$, which can detect only exact duplicates and cannot capture compositional similarity. Overall, these results indicate that the choice of normalization method affects the absolute uniqueness and novelty scores, but has limited impact on the relative model rankings, provided that the normalized distance $d$ does not increase too rapidly with $d'$. Since the motivation for introducing continuous metrics is to capture material similarity more granularly than binary metrics, it is important to choose a normalization method that preserves meaningful variation in $d'$. Our default normalization, $d = \frac{d'}{1+d'}$, appears to be a suitable choice for $d_\mathrm{elm}$ and $d_\mathrm{am}$, as its results are broadly consistent with those obtained using the other normalization methods. However, the optimal normalization strategy may depend on the distance function being normalized.

\subsection{Sensitivity Analysis for Coefficients of $d_\mathrm{elm+am}$}
\label{appendix:sec:sec:elm+am_coeff}

\begin{figure}
    \centering
    \includegraphics[width=\linewidth]{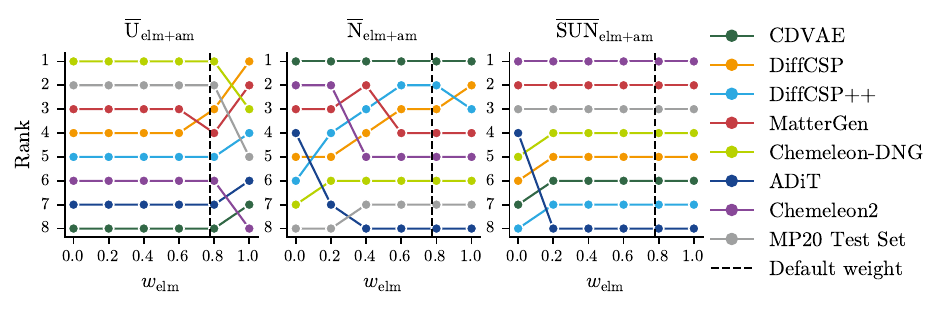}
    \caption{Model rankings for $\overline{\mathrm{U}}_\mathrm{elm+am}$, $\overline{\mathrm{N}}_\mathrm{elm+am}$, and $\overline{\mathrm{SUN}}_\mathrm{elm+am}$ under different weight settings for $d_\mathrm{elm+am}$. Increasing $w_\mathrm{elm}$ shifts the evaluation focus toward compositional uniqueness and novelty. The black dashed vertical line indicates the default weight used in this study, $w_\mathrm{elm} \approx 0.78$, which was statistically derived from the MP20 test set.}
    \label{appendix:fig:elm_am_weight_ranks}
\end{figure}

As shown in Equation~\ref*{eq:d_elm+am} in Section~\ref*{sec:sec:distance_definition} of the main text, $d_\mathrm{elm+am}$ is defined as a linear combination of $d_\mathrm{elm}$ and $d_\mathrm{am}$. The default weights, $w_\mathrm{elm} \approx 0.78$ and $w_\mathrm{am} \approx 0.22$, are chosen to balance the contributions of the two distance functions. However, these values should be regarded as a data-calibrated compromise rather than universal constants. Users who wish to place greater emphasis on either compositional or structural similarity may therefore choose alternative weights. In this section, we analyze how the model rankings change as these weights are varied.

Figure~\ref{appendix:fig:elm_am_weight_ranks} shows the model rankings based on $\overline{\mathrm{U}}_\mathrm{elm+am}$, $\overline{\mathrm{N}}_\mathrm{elm+am}$, and $\overline{\mathrm{SUN}}_\mathrm{elm+am}$, computed using different weights in $d_\mathrm{elm+am}$. The results show that the rankings for $\overline{\mathrm{U}}_\mathrm{elm+am}$ and $\overline{\mathrm{N}}_\mathrm{elm+am}$ change as the evaluation focus shifts from purely structural ($w_\mathrm{elm}=0.0$) to purely compositional ($w_\mathrm{elm}=1.0$). For example, the DiffCSP family, including DiffCSP and DiffCSP++, achieves higher uniqueness and novelty when compositional uniqueness and novelty are given greater weight. The consistently low uniqueness and high novelty of CDVAE indicate that its generated crystals are distant from the training samples but are not broadly dispersed, regardless of whether composition or structure is prioritized. In contrast, the model rankings for $\overline{\mathrm{SUN}}_\mathrm{elm+am}$ are largely insensitive to the coefficient values. This stability is likely attributable to the contribution of the stability term to $\overline{\mathrm{SUN}}_\mathrm{elm+am}$, which is independent of the coefficient values. For example, Chemeleon2, which ranks first in $\overline{\mathrm{SUN}}_\mathrm{elm+am}$ for all weight settings, generates more stable samples than the other models, as shown in Figure~\ref*{fig:stability_b_c} of the main text.

\subsection{Sensitivity Analysis for Continuous Stability Function}
\label{appendix:sec:sec:cs_sensitivity}

\begin{figure}
    \centering
    \includegraphics[width=0.7\linewidth]{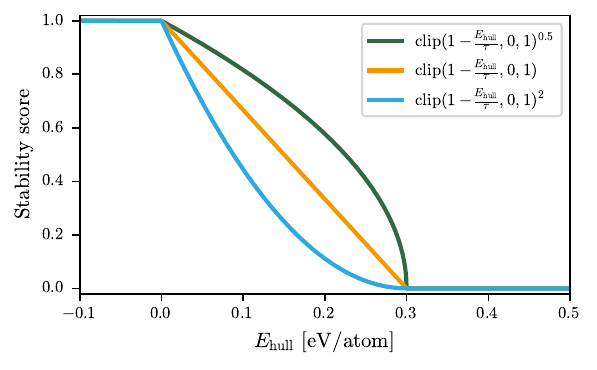}
    \caption{Visualizations of the three functional forms for continuous stability (cS), used in our analysis. The threshold hyperparameter $\tau$ is set to 0.3 [eV/atom].}
    \label{appendix:fig:cs_functional_form}
\end{figure}

By default, we define the continuous stability score, cS, using the simplest linear function, as shown in Equation~\ref*{eq:continuous-s} of the main text. We set the threshold $\tau$ to $0.4289$ [eV/atom], corresponding to the $99.9^\text{th}$ percentile of the DFT-derived $E_\mathrm{hull}$ distribution for the MP20 test set. This relatively loose threshold is used to allow as many generated samples as possible to receive a non-zero score. In the Materials Project database, approximately 4.9\% of experimentally synthesized structures have $E_\mathrm{hull}$ values above this threshold. If the threshold is set too low, many samples that may be synthesizable would make no contribution to the evaluation scores. Nevertheless, we acknowledge that both the functional form and the threshold value are hyperparameters. Therefore, this section investigates how the model rankings based on $\overline{\mathrm{SUN}}_\mathrm{elm+am}$ change when the functional form and threshold value are varied.

\begin{figure}
    \centering
    \includegraphics[width=\linewidth]{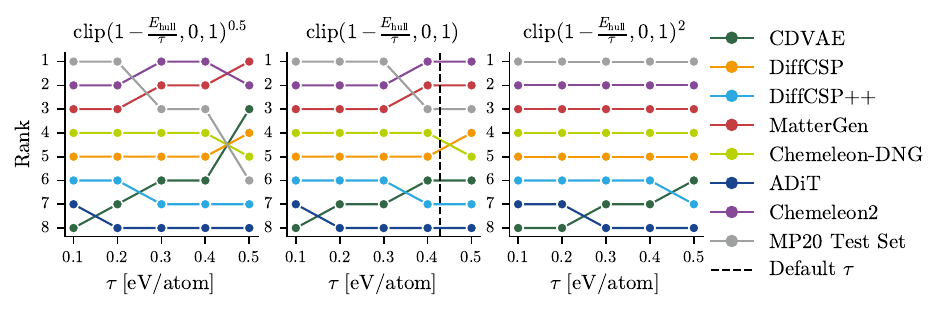}
    \caption{Model rankings for $\overline{\mathrm{SUN}}_\mathrm{elm+am}$ across different functional forms and thresholds for the continuous stability term. The black dashed vertical line marks the default stability function. Greater emphasis on stability, through a smaller $\tau$ or a larger exponent, increases the rank of the MP20 test set and decreases that of CDVAE. Apart from these two extreme cases, the remaining model rankings are largely stable across settings.}
    \label{appendix:fig:stability_shape_ranks}
\end{figure}

Figure~\ref{appendix:fig:cs_functional_form} shows the three functional forms compared in this analysis. The default form is $cS(x_i) = \mathrm{clip}\left(1 - \frac{E_\mathrm{hull}}{\tau}, 0, 1\right)$. Raising this function to the power of $0.5$ weakens the penalty, allowing samples with non-negligible $E_\mathrm{hull}$ values to contribute more to the score. In contrast, the squared form decreases more rapidly near $E_\mathrm{hull}=0$ and therefore penalizes such samples more strongly. Figure~\ref{appendix:fig:stability_shape_ranks} shows the model rankings based on $\overline{\mathrm{SUN}}_\mathrm{elm+am}$ under different functional forms and threshold values. As stability is given greater priority, that is, when $\tau$ is smaller or the exponent is larger, the MP20 test set ranks higher, whereas CDVAE ranks lower. This trend is consistent with their markedly different stability distributions shown in Figure~\ref*{fig:stability_b_c} of the main text. Excluding these two extreme cases, the rankings of the remaining models are largely consistent across functional forms and thresholds. Chemeleon2 and MatterGen are the two best-performing models under all settings. However, no model consistently outperforms the MP20 test set, as none produces a stability distribution comparable to that of the test set.

\subsection{Sensitivity Analysis for Weight Components of cSUN}
\label{appendix:sec:sec:csun_weight_sensitivity}

\begin{figure}
    \centering
    \includegraphics[width=\linewidth]{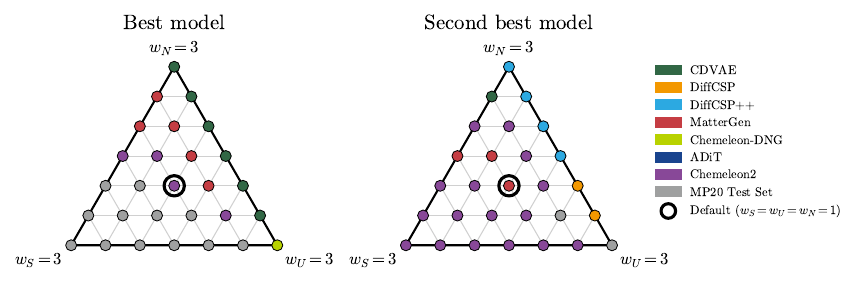}
    \caption{Models achieving the highest and second-highest $\overline{\mathrm{SUN}}_\mathrm{elm+am}$ scores under different component weights $(w_S, w_U, w_N)$. All weights are varied over $\{0.0, 0.5, 1.0, 1.5, 2.0, 2.5, 3.0\}$ while satisfying the constraint $w_S + w_U + w_N = 3$. The default setting is located at the center of each triangle. The rankings are most strongly influenced by $w_S$.}
    \label{appendix:fig:sun_component_weight_ternary}
\end{figure}

One advantage of the proposed cSUN metric is that it includes component weights, $w_S$, $w_U$, and $w_N$. These weights allow users to prioritize specific components over others. For example, Figure~\ref*{fig:sun_weight} in the main text shows how the model rankings based on $\overline{\mathrm{SUN}}$, defined using continuous distance functions, change as $w_S$ and $w_N$ are varied. This section provides a more detailed sensitivity analysis of these component weights.

Specifically, we vary all three weights over the set $\{0.0, 0.5, 1.0, 1.5, 2.0, 2.5, 3.0\}$ while maintaining the constraint $w_S + w_U + w_N = 3$. This range includes the default setting, $(w_S, w_U, w_N) = (1, 1, 1)$. Figure~\ref{appendix:fig:sun_component_weight_ternary} shows the best- and second-best-performing models on $\overline{\mathrm{SUN}}_\mathrm{elm+am}$ under each of the 28 weight settings. The results indicate that $w_S$ influences the model rankings most. When $w_S = 0$, corresponding to the right edge of the triangle where stability is ignored, CDVAE performs well. This is consistent with Figures~\ref*{fig:nov} and~\ref*{fig:nov_samples} in the main text, which show that CDVAE achieves high novelty by generating physically implausible samples. Once stability is considered, Chemeleon2 and MatterGen outperform the other generative models under most settings, confirming their robustness across a broad range of evaluation priorities. When stability is sufficiently prioritized ($w_S \ge 2$), the MP20 test set consistently achieves the best performance.

\subsection{Distribution of cU and cN Scores}
\label{appendix:sec:sec:dist_cu_cn_elm+am}

\begin{figure}
    \centering
    \includegraphics[width=\linewidth]{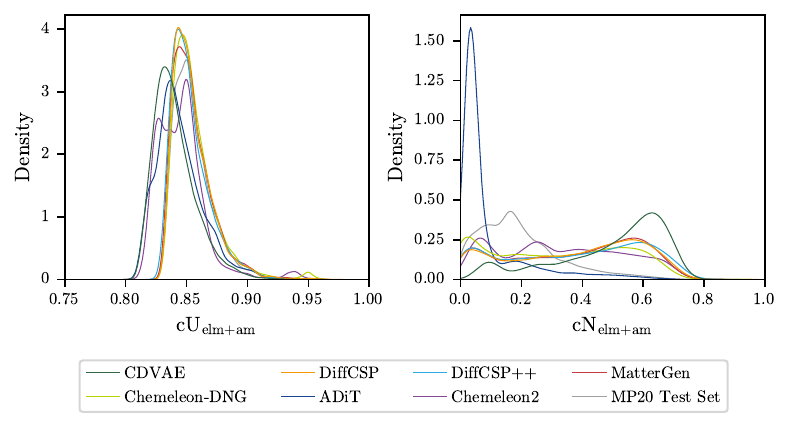}
    \caption{Distributions of continuous uniqueness $\mathrm{cU}_\mathrm{elm+am}$ and novelty $\mathrm{cN}_\mathrm{elm+am}$ scores for 10k samples generated by different models.}
    \label{appendix:fig:dist_cu_cn_elm+am}
\end{figure}

Section~\ref*{sec:sec:un_experiment} (main text) compares aggregated $\overline{\mathrm{U}}$ and $\overline{\mathrm{N}}$ scores across models. To obtain a more detailed view, we examine the sample-level distributions of cU and cN. Figure~\ref{appendix:fig:dist_cu_cn_elm+am} shows the $\mathrm{cU}_\mathrm{elm+am}$ and $\mathrm{cN}_\mathrm{elm+am}$ distributions for each model and the MP20 test set baseline. These distributions reveal information that is not captured by the aggregated scores in Figures~\ref*{fig:uni} and~\ref*{fig:nov} in the main text. For uniqueness, DiffCSP, DiffCSP++, MatterGen, and Chemeleon-DNG show distributions that closely resemble the MP20 test set, suggesting that their generated samples are scattered through material space in a similar manner. The small peak near $\mathrm{cU}_\mathrm{elm+am}=0.95$ likely contributes to the highest aggregated uniqueness score of Chemeleon-DNG. In contrast, CDVAE, ADiT, and Chemeleon2 underperform relative to the baseline, despite a small peak near $0.94$ for Chemeleon2. For novelty, the model-dependent differences are more substantial. CDVAE’s high novelty is likely explained by its tendency to generate physically implausible structures, as discussed in Figure~\ref*{fig:nov_samples} (main text). ADiT, by contrast, shows much lower novelty than the other models, consistent with its very limited ability to generate new compositions in the $\overline{\mathrm{N}}_\mathrm{comp}$ and $\overline{\mathrm{N}}_\mathrm{elm}$ subplots of Figure~\ref*{fig:nov} (main text). DiffCSP, DiffCSP++, MatterGen, and Chemeleon-DNG again show highly similar distributions, but with novelty higher than the MP20 baseline. The similarity between these four models may stem from their similar architecture (equivariant graph neural networks) and training objective (diffusion in the space of lattice parameters, atomic species, and fractional coordinates).

\subsection{Distance Between Distributions of Different Models}
\label{appendix:sec:sec:model_dist}

\begin{figure}
    \centering
    \includegraphics[width=\linewidth]{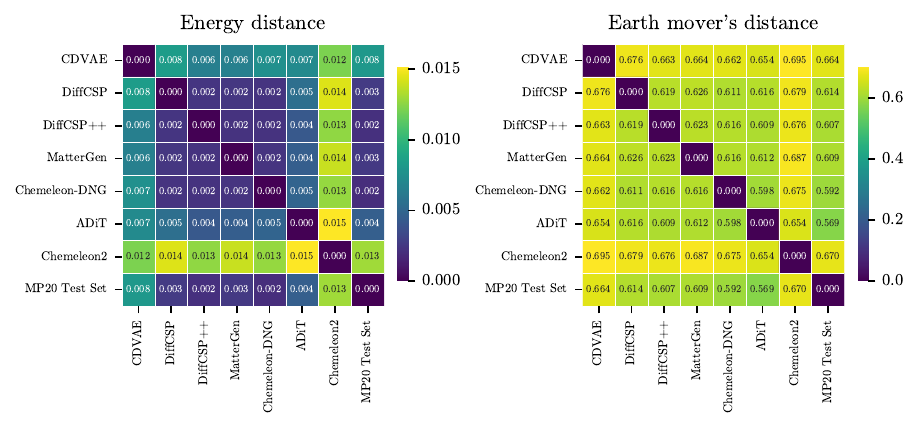}
    \caption{Pairwise energy distance and earth mover’s distance between 1,000-sample sets from each generative model and the MP20 test set. $d_\mathrm{elm+am}$ is used as the underlying distance function for both metrics.}
    \label{appendix:fig:model_dist}
\end{figure}

This study evaluates uniqueness, novelty, and stability, each of which uses a different reference set. Uniqueness is computed within the generated set itself, novelty is measured relative to the training set, and stability is evaluated against the reference database used to construct the convex hull. A further comparison, not considered in these metrics, is between the generated distributions of different models. Here, we therefore compare the distributions of different generative models.

We quantify the distance between two generated sample sets using the energy distance and the earth mover’s distance:
\begin{align*}
\mathrm{Energy \ distance} &\coloneqq \frac{2}{n^2} \sum_{i, j} d(x_i, y_j) - \frac{1}{n^2} \sum_{i, i'} d(x_i, x_{i'}) - \frac{1}{n^2} \sum_{j, j'} d(y_j, y_{j'}),\\
\mathrm{Earth \ mover's \ distance} &\coloneqq \frac{1}{n} \min_{\sigma} \sum_i d(x_i, y_{\sigma(i)}),
\end{align*}
where $\{x_i\}_{i=1}^n$ and $\{y_j\}_{j=1}^n$ are samples generated by two different models, and $\sigma$ is a one-to-one mapping between them. In both cases, smaller values indicate more similar distributions, and identical empirical distributions yield a distance of zero. It should be noted that the energy distance may, in principle, take a negative value, although this did not occur in our experiments. Figure~\ref{appendix:fig:model_dist} reports both distances for all pairs of generative models and the MP20 test set. We use $d_\mathrm{elm+am}$ as the underlying distance function. Chemeleon2 is clearly separated from the other models and from the MP20 baseline under both metrics, likely reflecting the effect of RL on its generated distribution. CDVAE shows the next most distinctive distribution and approximates the MP20 distribution less accurately than the remaining models, except for Chemeleon2. This may be because CDVAE is an early generative model with limited capacity to capture the full data distribution. The energy distances among DiffCSP, DiffCSP++, MatterGen, and Chemeleon-DNG are consistently smaller than those for other model pairs, possibly owing to their similar architectures and objectives: equivariant graph neural networks trained by diffusion in the space of lattice parameters, atomic species, and fractional coordinates. This pattern, however, is not observed for the earth mover’s distance.

\bibliographystylesi{iopart-num-long}
\bibliographysi{reference}

\end{document}